\documentclass{article}

\usepackage[final]{nips_2018} % produce camera-ready copy

\usepackage[utf8]{inputenc} % allow utf-8 input
\usepackage[T1]{fontenc}    % use 8-bit T1 fonts
\usepackage{hyperref}       % hyperlinks
\usepackage{url}            % simple URL typesetting
\usepackage{color} 
\usepackage{amsfonts}       % blackboard math symbols
\usepackage{nicefrac}       % compact symbols for 1/2, etc.
\usepackage{microtype}      % microtypography

\usepackage{microtype}
\usepackage{graphicx}
\usepackage{wrapfig}
\usepackage{booktabs} % for professional tables

\usepackage{hyperref}

\usepackage{natbib}
\usepackage{algorithm}
\usepackage{algorithmic}
\usepackage{balance}
\usepackage{amsmath}
\usepackage{amsthm}
\usepackage{amssymb}
\usepackage{adjustbox}
\usepackage{bm}
\usepackage{url}
\usepackage{stmaryrd}
\usepackage{multirow}
\usepackage{epsfig} 
\usepackage{epstopdf}
\usepackage{booktabs}
\usepackage{pgffor}
\usepackage{flexisym}  % \textprime which can be directly used in math or in text mode
\usepackage{cases}
\usepackage{enumitem}
\DeclareMathOperator*{\argmin}{\arg\min}
\DeclareMathOperator*{\argmax}{\arg\max}

\newtheorem{exe}{Theorem}
\newtheorem{definition}{Definition}
\newtheorem{theorem}{Theorem}

\newtheorem{lemma}[theorem]{Lemma}
\renewcommand{\algorithmicrequire}{\textbf{Input:}}
\renewcommand{\algorithmicensure}{\textbf{Output:}}

\makeatletter
\def\@xfootnote[#1]{%
  \protected@xdef\@thefnmark{#1}%
  \@footnotemark\@footnotetext}
\makeatother

\usepackage{graphicx}
\usepackage{wrapfig}

\usepackage{footnote}
\makesavenoteenv{tabular}
\makesavenoteenv{table}

\usepackage[english]{babel}
\usepackage{lineno, blindtext}

\newcommand{\ours}[0]{\text{SURF}}
\title{Boosted Sparse and Low-Rank Tensor Regression}

\author{
  Lifang~He \\
  Weill~Cornell~Medicine\\
  \texttt{lifanghescut@gmail.com}\\
  \And
  Kun~Chen\thanks{Corresponding Author} \\
  University~of~Connecticut \\
  \texttt{kun.chen@uconn.edu} \\
  \And
  Wanwan~Xu \\
  University~of~Connecticut \\
  \texttt{wanwan.xu@uconn.edu} \\
  \And
  Jiayu~Zhou \\
  Michigan~State~Universtiy \\
  \texttt{dearjiayu@gmail.com} \\
  \And
  Fei~Wang \\
  Weill~Cornell~Medicine\\
  \texttt{few2001@med.cornell.edu} \\
}

\begin{document}

\maketitle

\begin{abstract}
We propose a sparse and low-rank tensor regression model to relate a univariate outcome to a feature tensor, in which each unit-rank tensor from the CP decomposition of the coefficient tensor is assumed to be sparse. This structure is both parsimonious and highly interpretable, as it implies that the outcome is related to the features through a few distinct pathways, each of which may only involve subsets of feature dimensions. We take a divide-and-conquer strategy to simplify the task into a set of sparse unit-rank tensor regression problems. To make the computation efficient and scalable, for the unit-rank tensor regression, we propose a stagewise estimation procedure to efficiently trace out its entire solution path. We show that as the step size goes to zero, the stagewise solution paths converge exactly to those of the corresponding regularized regression. The superior performance of our approach is demonstrated on various real-world and synthetic examples.  

% {\color{red} (Lifang, how about this paragraph? You may further modify.)}

%Sparse learning is effective when both prediction and interpretation are needed. The LASSO and its extensions are standard methods for selecting a subset of features while optimizing a prediction function. In this paper, we are interested in sparse and low-rank tensor regression, wherein (1) there is no need to reshape the multidimensional data into vectors and (2) structural information from multiple dimensions are taken into account. Our main contribution is a new method called boosted sparse and low-rank tensor regression (BSLTR) for automatically selecting a subset of features while optimizing prediction for multidimensional data.
% Abstracts must be a single paragraph, ideally between 4--6 sentences long.
\end{abstract}

\vspace{-5pt}
\section{Introduction}
\label{sect:intro}
\vspace{-2pt}
Regression analysis is commonly used for modeling the relationship between a predictor vector $\mathbf{x} \in \mathbb{R}^I$ and a scalar response $y$. Typically a good regression model can achieve two goals: accurate prediction on future response and parsimonious interpretation of the dependence structure between $y$ and $\mathbf{x}$ \citep{hastie2009elements}. As a general setup, it fits $M$ training samples $\{(\mathbf{x}^m, y^m)\}_{m=1}^M$ via minimizing a regularized loss, i.e., a loss $L(\cdot)$ plus a regularization term $\Omega(\cdot)$, as follows 
\begin{align}
  \underset{\mathbf{w}}{\min} \frac{1}{M}\sum\nolimits_{m=1}^M L(\langle \mathbf{x}^m, \mathbf{w} \rangle, y^m) + \lambda \Omega(\mathbf{w}),
  \label{eq:LR}
\end{align}
where $\mathbf{w} \in \mathbb{R}^{I}$ is the regression coefficient vector, $\langle \cdot, \cdot \rangle$ is the standard Euclidean inner product, and $\lambda>0$ is the regularization parameter. For example, the sum of squared loss with $\ell_1$-norm regularization leads to the celebrated LASSO approach \citep{tibshirani1996regression}, which performs sparse estimation of $\mathbf{w}$ and thus has implicit feature selection embedded therein.

In many modern real-world applications, the predictors/features are represented more naturally as higher-order tensors, such as videos and Magnetic Resonance Imaging (MRI) scans. In this case, if we want to predict a response variable for each tensor, a naive approach is to perform linear regression on the vectorized data (e.g., by stretching the tensor element by element). However, it completely ignores the multidimensional structure of the tensor data, such as the spatial coherence of the voxels. This motivates the tensor regression framework \citep{yu2016learning,zhou2013tensor}, which treats each observation as a tensor $\mathcal{X}$ and learns a tensor coefficient $\mathcal{W}$ via regularized model fitting:
\begin{align}
  \underset{\mathcal{W}}{\min} \frac{1}{M} \sum\nolimits_{m=1}^M L(\langle \mathcal{X}^m, \mathcal{W} \rangle, y^m) + \lambda \Omega(\mathcal{W}).
  \label{eq:MLR}
\end{align}
When $\ell_1$-norm regularization is used, this formulation is essentially equivalent to (\ref{eq:LR}) via vectorization. To effectively exploit the structural information of $\mathcal{X}^m$, we can impose a low-rank constraint on $\mathcal{W}$ for Problem (\ref{eq:MLR}). Some authors achieved this by fixing the CANDECOMP/PARAFAC (CP) rank of $\mathcal{W}$ as a priori. For example, \citep{su2012multivariate} assumed $\mathcal{W}$ to be rank-1. Since rank-1 constraint is too restrictive, \citep{guo2012tensor} and \citep{zhou2013tensor} imposed a rank-$R$ constraint in a tensor decomposition model, but none of the above methods considered adding sparsity constraint as well to enhance the model interpretability. \citep{wang2014clinical} imposed a restrictive rank-1 constraint on $\mathcal{W}$ and also applied an elastic net regularization \citep{zou2005regularization}. \citep{tan2012logistic} imposed a rank-$R$ constraint and applied $\ell_1$-norm regularization to factor matrices to also promote sparsity in $\mathcal{W}$. \citep{signoretto2014learning} applied the trace norm (nuclear norm) for low-rank estimation of $\mathcal{W}$, and \citep{song2017multilinear} imposed a combination of trace norm and $\ell_1$-norm on $\mathcal{W}$. \citep{bengua2017efficient} showed that the trace norm may not be appropriate for capturing the global correlation of a tensor as it provides only the mean of the correlation between a single mode (rather than a few modes) and the rest of the tensor. In all the above sparse and low-rank tensor models, the sparsity is imposed on $\mathcal{W}$ itself, which, does not necessarily lead to the sparsity on the decomposed matrices.

In this paper, we propose a sparse and low-rank tensor regression model in which the unit-rank tensors from the CP decomposition of the coefficient tensor are assumed to be sparse. This structure is both parsimonious and highly interpretable, as it implies that the outcome is related to the features through a few distinct pathways, each of which may only involve subsets of feature dimensions. We take a divide-and-conquer strategy to simplify the task into a set of sparse unit-rank tensor factorization/regression problems (SURF) in the form of
\begin{align*}
  \underset{\mathcal{W}}{\min} \frac{1}{M} \sum\nolimits_{m=1}^M(y^m-\langle\mathcal{X}^m,\mathcal{W}\rangle)^2 + \lambda \|\mathcal{W}\|_1, \quad \text{s.t.} \quad \text{rank}(\mathcal{W}) \leq 1,
\end{align*}
To make the solution process efficient for the SURF problem, we propose a boosted/stagewise estimation procedure to efficiently trace out its entire solution path. We show that as the step size goes to zero, the stagewise solution paths converge exactly to those of the corresponding regularized regression. The effectiveness and efficiency of our proposed approach is demonstrated on various real-world datasets as well as under various simulation setups.

%lead to good interpretation because the sparsity of $\mathcal{W}$ does not necessarily 

% To the best of knowledge, none of the previous works has investigated the problem of regularization parameter selection in this context so far.
% Nonetheless, these regression methods are formulated by fixing the rank in advance, and they do not XXX either.
\vspace{-5pt}
\section{Preliminaries on Tensors}
\vspace{-5pt}
\label{sect:prelim}
% In this section, we briefly introduce some notations and concepts of tensor algebra.
% Table \ref{tab:notation} lists basic symbols that will be used throughout the paper.
% We give the notation and preliminaries from tensor algebra \citep{kolda2009tensor}. 
% We begin with the notation and background necessary for the sequel. 
% Tensors are mathematical objects for representing multidimensional arrays; vectors and matrices are first-order and second-order special cases of tensors, respectively. To facilitate the distinction
% The order of a tensor is the number of dimensions, also known as ways or modes.
We start with a brief review of some necessary preliminaries on tensors, and more details can be found in \citep{kolda2009tensor}. We denote scalars by lowercase letters, e.g., $a$; vectors by boldfaced lowercase letters, e.g., $\mathbf{a}$; matrices by boldface uppercase letters, e.g., $\mathbf{A}$; and tensors by calligraphic letters, e.g., $\mathcal{A}$. We denote their entries by $a_i$, $a_{i,j}$, $a_{i,j,k}$, etc., depending on the number of dimensions. Indices are denoted by lowercase letters spanning the range from 1 to the uppercase letter of the index, e.g., $n= 1, \cdots, N$. Each entry of an $N$th-order tensor $\mathcal{A} \in \mathbb{R}^{I_1 \times \cdots \times I_N}$ is indexed by $N$ indices $\{i_n\}_{n=1}^N$, and each $i_n$ indexes the $n$-mode of $\mathcal{A}$. Specifically, $-n$ denotes every mode except $n$.

% All vectors are column vectors unless otherwise specified.

% Unfolding $\mathcal{A}$ along the $n$-mode is denoted as $\mathbf{A}_{(n)} \in \mathbb{R}^{I_n \times (I_1 \times \cdots \times I_{n-1} \times I_{n+1} \times \cdots \times I_N)}$, where the column vectors of $\mathbf{A}_{(n)}$ are the $n$-mode vectors of $\mathcal{A}$.

\begin{definition}[Inner Product]
The inner product of two tensors $\mathcal{A}, \mathcal{B} \in \mathbb{R}^{I_1\times\cdots\times I_N}$ is the sum of the products of their entries, defined as $\langle \mathcal{A},\mathcal{B} \rangle = \sum_{i_1=1}^{I_1} \cdots \sum_{i_1=1}^{I_N} a_{i_1, \cdots, i_N} b_{i_1, \cdots, i_N}$.
% \begin{equation}
%     \langle \mathcal{A},\mathcal{B} \rangle = \sum_{i_1=1}^{I_1} \cdots \sum_{i_1=1}^{I_N} a_{i_1, \cdots, i_N} b_{i_1, \cdots, i_N}.
% \end{equation}
\end{definition}
\vspace{-2pt}
It follows immediately that the Frobenius norm of $\mathcal{A}$ is defined as $\|\mathcal{A}\|_F = \sqrt{\langle \mathcal{A},\mathcal{A} \rangle}$. The $\ell_1$ norm of a tensor is defined as $\|\mathcal{A}\|_1 = \sum_{i_1=1}^{I_1}\cdots\sum_{i_N=1}^{I_N} |\mathcal{A}_{i_1, \cdots, i_N}|$.
% \begin{definition}[$\ell_1$-Norm]
% The $\ell_1$-norm of a tensor $\mathcal{A} \in \mathbb{R}^{I_1\times\cdots\times I_N}$ is defined as 
% \begin{equation}
%     \| \mathcal{A} \|_1 = \sum_{i_1=1}^{I_1}\sum_{i_2=1}^{I_2}\cdots\sum_{i_N=1}^{I_N}|a_{i_1, i_2, \cdots, i_N}|.  
% \end{equation}
% \end{definition}
\begin{definition}[Tensor Product]
Let $\mathbf{a}^{(n)} \in \mathbb{R}^{I_n}$ be a length-$I_n$ vector for each $n = 1, 2, \cdots, N$. The tensor product of $\{\mathbf{a}^{(n)}\}$, denoted by $\mathcal{A} = \mathbf{a}^{(1)} \circ \cdots \circ \mathbf{a}^{(N)}$, is an $(I_1 \times \cdots \times I_N)$-tensor of which the entries are given by $\mathcal{A}_{i_1, \cdots, i_N} = a_{i_1}^{(1)} \cdots a_{i_N}^{(N)}$.
% \begin{equation}
%     a_{i_1, \cdots, i_N} = a_{i_1}^{(1)} \cdots a_{i_N}^{(N)}.
% \end{equation}
We call $\mathcal{A}$ a rank-1 tensor or a unit-rank tensor.
\end{definition}
% In particular, for $\mathcal{A}=\mathbf{a}^{(1)} \circ \cdots \circ \mathbf{a}^{(N)}$ and $\mathcal{B}=\mathbf{b}^{(1)} \circ \cdots \circ \mathbf{b}^{(N)}$, it holds that
% \begin{equation}
% \big \langle \mathcal{A},\mathcal{B} \big \rangle = \prod_{n=1}^{N} \big \langle \mathbf{a}^{(n)}, \mathbf{b}^{(n)} \big \rangle = \prod_{n=1}^{N} {\mathbf{a}^{(n)}}^{\mathrm{T}} \mathbf{b}^{(n)}.
% \label{eq:inner_rank1}    
% \end{equation}

% Different from matrices, there are multiple ways to define the rank of a tensor \citep{kolda2009tensor}. Here, we define the tensor rank based on the CP decomposition. 

\begin{definition}[CP Decomposition]
Every tensor $\mathcal{A} \in \mathbb{R}^{I_1\times\cdots\times I_N}$ can be decomposed as a weighted sum of rank-1 tensors with a suitably large $R$ as  
\begin{align}
\mathcal{A} = \sum\nolimits_{r=1}^{R} \sigma_r \cdot \mathbf{a}_{r}^{(1)} \circ \cdots \circ \mathbf{a}_{r}^{(N)},
\label{eq:CP}
\end{align}
where $\sigma_r \in \mathbb{R}$, $\mathbf{a}_{r}^{(n)} \in \mathbb{R}^I$ and $\|\mathbf{a}_{r}^{(n)}\|_2 = 1$. 
% for $n = 1, 2, \cdots, N$, $\mathbf{A}^{(n)}= [\mathbf{a}_{1}^{(n)}, \cdots, \mathbf{a}_{K}^{(n)} ]$ are factor matrices of size $I_n \times K$ with unit column norms. $K$ is the number of factors, and $\llbracket \cdot \rrbracket$ is used for shorthand.
\end{definition}

%The CP decomposition is also known as the rank-$K$ decomposition, and the matrices $\mathbf{A}^{(k)}=[\mathbf{a}_k^{(1)}, \cdots, \mathbf{a_}k^{(N)}]$ for $k=1, \cdots, K$ are called factor matrices.

\begin{definition}[Tensor Rank]
The tensor rank of $\mathcal{A}$, denoted by $rank(\mathcal{A})$, is the smallest number $R$ such that the equality (\ref{eq:CP}) holds.
\end{definition}
%In general, there is no closed-form solution to determine the rank of a tensor of order three or higher a priori, and there is no straightforward algorithm to compute the tensor rank.
%In practice, one often guesses a value for $R$, and searches for the factor matrices that best fit $\mathcal{A}$. The best fit measure is often the Frobenius norm.

% \begin{definition}[$n$-mode Product]
% The $n$-mode product of a tensor $\mathcal{A}\in \mathbb{R}^{I_{1}\times\cdots \times I_N}$ by a matrix $\mathbf{U}\in \mathbb{R}^{J_n \times I_n}$, denoted by $\mathcal{A} \times_{n}\mathbf{U}$, is an $(I_1 \times \cdots \times I_{n-1} \times J_n \times I_{n+1} \cdots \times I_N)$-tensor of which the entries are given by
% \begin{equation}
% (\mathcal{A}\times_{n}\mathbf{U})_{i_1,\dots, i_{n-1},j_n,i_{n+1},\dots, i_N} = \sum_{i_n=1}^{I_n}a_{i_1, \dots, i_N}u_{j_n,i_n}.
% \end{equation}
% \end{definition}
\begin{definition}[$n$-mode Product]
The $n$-mode product of a tensor $\mathcal{A}\in \mathbb{R}^{I_{1}\times\cdots \times I_N}$ by a vector $\mathbf{u}\in \mathbb{R}^{I_n}$, denoted by $\mathcal{A} \times_{n}\mathbf{u}$, is an $(I_1 \times \cdots \times I_{n-1} \times I_{n+1} \cdots \times I_N)$-tensor of which the entries are given by $(\mathcal{A}\times_{n}\mathbf{u})_{i_1,\dots, i_{n-1}i_{n+1},\dots, i_N} = \sum_{i_n=1}^{I_n}a_{i_1, \dots, i_N}u_{i_n}.$
% \begin{equation}
% (\mathcal{A}\times_{n}\mathbf{u})_{i_1,\dots, i_{n-1}i_{n+1},\dots, i_N} = \sum_{i_n=1}^{I_n}a_{i_1, \dots, i_N}u_{i_n}. \nonumber
% \end{equation}
\end{definition}
% In practice, one often prefers to fit a multi-linear model of lower rank, F < rank{T}, fixed in advance, so that we have to deal with an approximation problem. More precisely, it is aimed at minimizing an objective function of the form

% Note that for $\mathcal{A}=\mathbf{a}^{(1)} \circ \cdots \circ \mathbf{a}^{(N)}$ and $\mathcal{B}=\mathbf{b}^{(1)} \circ \cdots \circ \mathbf{b}^{(N)}$, it holds that
% \begin{align}
% \langle \mathcal{A},\mathcal{B} \rangle = \prod_{n=1}^{N} \langle \mathbf{a}^{(n)}, \mathbf{b}^{(n)} \rangle = \prod_{n=1}^{N} {\mathbf{a}^{(n)}}^{\mathrm{T}} \mathbf{b}^{(n)}~.
% \label{eq:inner_rank1}
% \end{align}

% \input{table_notation}
\section{Sparse and Low-Rank Tensor Regression}
\label{sect:method}
% We introduce the details of {\ours} in this section. First we will formally define the problem to establish the learning objective.
% \subsection{Model}
\vspace{-3pt}
\subsection{Model Formulation}
\vspace{-3pt}
For an $N$th-order predictor tensor  $\mathcal{X}^m \in \mathbb{R}^{I_1 \times \cdots \times I_N}$ and a scalar response $y^m$, $m=1, \cdots, M$, we consider the regression model of the form
\begin{equation}
    y^m = \langle \mathcal{X}^m, \mathcal{W} \rangle + \varepsilon^m
\end{equation}
where $\mathcal{W} \in \mathbb{R}^{I_1 \times \cdots \times I_N}$ is an $N$th-order coefficient tensor, and $\varepsilon^m$ is a random error term of zero mean. Without loss of generality, the intercept is set to zero by centering the response and standardizing the predictors as $\sum_{m=1}^M y^m = 0$; $\sum_{m=1}^M \mathcal{X}_{i_1, \cdots, i_N}^m = 0$ and $\sum_{m=1}^M (\mathcal{X}_{i_1, \cdots, i_N}^m)^2/M = 1$ for $i_n = 1, \cdots, I_n$. 
Our goal is to estimate $\mathcal{W}$ with $M$ i.i.d. observations $\{(\mathcal{X}^m, y^m)\}_{m=1}^M$.

% To facilitate estimation of high-dimensional coefficient tensor $\mathcal{W}$ 
To reduce the complexity of the model and leverage the structural information in $\mathcal{X}^m$, we assume that the coefficient tensor $\mathcal{W}$ to be both low-rank and sparse. Specifically, we assume $\mathcal{W}$ can be decomposed via CP decomposition as
%\begin{align*}
    $\mathcal{W} =\sum_{r=1}^{R}\mathcal{W}_r =\sum_{r=1}^{R} \sigma_r \mathbf{w}_{r}^{(1)} \circ \cdots \circ \mathbf{w}_{r}^{(N)}$,
%\end{align*}
where each rank-1 tensor $\mathcal{W}_r$ is possibly sparse, or equivalently, the vectors in its representation $\mathbf{w}_{r}^{(n)}$, $n= 1,\ldots, N$, are possibly sparse, for all $r = 1,\ldots, R$.

Here we impose sparsity on the rank-1 components from CP decomposition -- rather than on $\mathcal{W}$ itself \citep{chen2012reduced, tan2012logistic}. This adaption can be more beneficial in multiple ways: 1) It integrates a finer sparsity structure into the CP decomposition, which enables a direct control of component-wise sparsity; 2) It leads to an appealing model interpretation and feature grouping: the outcome is related to the features through a few distinct pathways, each of which may only involve subsets of feature dimensions; 3) It leads to a more flexible and parsimonious model as it requires less number of parameters to recover the within-decomposition sparsity of a tensor than existing methods which impose sparsity on the tensor itself, thus makes the model generalizability better.

A straightforward way of conducting model estimation is to solve the following optimization problem:
\begin{align}
& \underset{\sigma_r, \mathbf{w}_r^{(n)}}{\min} \frac{1}{M} \sum\nolimits_{m=1}^M(y^m - \langle \mathcal{X}^m, \sum\nolimits_{r=1}^{R} \sigma_r \mathbf{w}_{r}^{(1)} \circ \cdots \circ \mathbf{w}_{r}^{(N)} \rangle)^2 +  
\sum\nolimits_{r=1}^R \sum\nolimits_{n =1}^{N} \lambda_{r,n} \|\mathbf{w}_{r}^{(n)}\|_1,  \nonumber \\
%\sum_{r=1}^R \lambda_r \|\sigma_r \mathbf{w}_{r}^{(1)} \circ \cdots \circ \mathbf{w}_{r}^{(N)}\|_1,  \nonumber \\
&\quad \text{s.t.} \quad \| \mathbf{w}_{r}^{(n)} \|_1 = 1, ~n=1,\cdots, N, ~ r=1, \cdots, R.
  \label{eq:MLR_rank_R}
\end{align}
where $\lambda_{r,n}$, $r=1,\cdots, R, n=1,\cdots,N$, are regularization parameters. This problem is very difficult to solve because: 1) The CP rank $R$ needs to be pre-specified; 2) As the CP decomposition may not be unique, the pursue of its within-decomposition sparsity is highly non-convex and the problem may suffer from parameter identifiability issues \citep{mishra2017sequential}; 3) The estimation may involve many regularization parameters, for which the tuning becomes very costly.

\vspace{-2pt}
\subsection{Divide-and-Conquer: Sequential Pursue for Sparse Tensor Decomposition}
\vspace{-2pt}
%There are two different approaches for solving the problem (\ref{eq:MLR_rank_R_matrix_Enet}): simultaneous and sequential methods \cite{zhao2013higher}. In a simultaneous method, all factors are calculated simultaneously by minimizing a certain criterion. 
We propose a divide-and-conquer strategy to recover the sparse CP decomposition. Our approach is based on the sequential extraction method (a.k.a. deflation) \citep{phan2015tensor,mishra2017sequential}, which seeks a unit-rank tensor at a time and then deflates to find further unit-rank tensors from the residuals. This has been proved to be a rapid and effective method of partitioning and concentrating tensor decomposition.
% which has been widely used in tensor partial least squares regression \citep{westerhuis2001deflation,zhao2013higher}.
% The standard sequential extraction of a rank-1 tensor has been used in practice partial least squares tensor regression
Specifically, we sequentially solve the following sparse unit-rank problem:
%\begin{align}
%  \widehat{\mathcal{W}}_{r} = \underset{\mathcal{W}_r}{\min} \frac{1}{M} \sum_{m=1}^M(y^m_r -\langle\mathcal{X}^m,\mathcal{W}_r\rangle)^2 + \lambda_r \|\mathcal{W}_r\|_1, \quad \text{s.t.} \quad \text{rank}(\mathcal{W}_r) \leq 1,
%  \label{eq:MLR_rank_1_matrix_LASSO}
%\end{align}
\begin{align}
  \widehat{\mathcal{W}}_{r} = \underset{\mathcal{W}_r}{\min} \frac{1}{M} \sum\nolimits_{m=1}^M(y^m_r -\langle\mathcal{X}^m,\mathcal{W}_r\rangle)^2 + \lambda_r \|\mathcal{W}_r\|_1 + \alpha \|\mathcal{W}_r\|_F^2, \quad \text{s.t.} \quad \text{rank}(\mathcal{W}_r) \leq 1.
  \label{eq:MLR_rank_1_matrix_Enet}
\end{align}
where $r$ is the sequential number of the unit-rank terms and $y_r^m$ is the current residue of response with
\[
y^m_r :=
\begin{cases}
y^m, & \text{if}~~r=1 \\
y^m_{r-1} - \langle \mathcal{X}^m, \widehat{\mathcal{W}}_{r-1} \rangle, & \text{otherwise}, \nonumber
\end{cases}
\]
where $\widehat{\mathcal{W}}_{r-1}$ is the estimated unit-rank tensor in the $(r-1)$-th step, with tuning done by, e.g., cross validation. The final estimator is then obtained as $\widehat{\mathcal{W}}(R) = \sum_{r=1}^R \widehat{\mathcal{W}}_{r}$. Here for improving the convexity of the problem and its numerical stability, we have used the elastic net \citep{zou2005regularization} penalty form instead of LASSO, which is critical to ensure the convergence of the optimization solution. The accuracy of the solution can be controlled simply by adjusting the values of $\lambda_r$ and $\alpha$. Since we mainly focus on sparse estimation, we fix $\alpha>0$ as a small constant in numerical studies.

As each $\mathcal{W}_{r}$ is of unit rank, it can be decomposed as $\mathcal{W}_{r} = \sigma_r \mathbf{w}^{(1)}_r \circ \cdots \circ \mathbf{w}^{(N)}_r$ with $\sigma_r\geq 0$ and $\|\mathbf{w}^{(n)}_r\|_1 = 1, n=1, \cdots, N$. It is then clear that $\|\mathcal{W}_{r}\|_1=\|\widehat{\sigma}_r \widehat{\mathbf{w}}^{(1)}_r \circ \cdots \circ \widehat{\mathbf{w}}^{(N)}_r\|_1 = \sigma_r \prod_{n = 1}^N \|\mathbf{w}_r^{(n)}\|_1$. That is, the sparsity of a unit-rank tensor directly leads to the sparsity of its components. This allows us to kill multiple birds with one stone: by simply pursuing the element-wise sparsity of the unit-rank coefficient tensor with only one tuning parameter $\lambda$, solving (\ref{eq:MLR_rank_1_matrix_Enet}) can produce a set of sparse factor coefficients $\widehat{\mathbf{w}}^{(n)}_r$ for $n=1, \cdots, N$ simultaneously. 

With this sequential pursue strategy, the general problem boils down to a set of sparse unit-rank estimation problems, for which we develop a novel stagewise/boosting algorithm. 
\vspace{-5pt}
\vspace{-5pt}
\section{Fast Stagewise Unit-Rank Tensor Factorization (SURF)}
\vspace{-5pt}
% The most straightforward solution to (\ref{eq:MLR_object_k}) is arguably alternating least squares (ACS) or block-coordinate based methods. However, ...
% Here we propose an efficient iterative algorithm to solve (\ref{eq:MLR_rank_R_matrix}), by means of a sequential greedy stagewise estimation along $K$ rank-1 terms.
% The advantage of this algorithm over the standard optimization algorithms for tensor problems is that it computes solutions along the entire ``regularization path'' (the set of solutions for all values of $\lambda$), by only using the differences of the loss function. Therefore, it is applicable to loss functions that are not differentiable or of which differentiation is too complex or computationally expensive.
For simplicity, we drop the index $r$ and write the generic form of the problem in (\ref{eq:MLR_rank_1_matrix_Enet}) as
\begin{align}
  \widehat{\mathcal{W}} = \underset{\mathcal{W}}{\min} \frac{1}{M} \sum\nolimits_{m=1}^M(y^m -\langle\mathcal{X}^m,\mathcal{W}\rangle)^2 + \lambda \|\mathcal{W}\|_1 + \alpha \|\mathcal{W}\|_F^2, \quad \text{s.t.} \quad \text{rank}(\mathcal{W}) \leq 1.
  \label{eq:MLR_SURT}
\end{align}
Let $\mathcal{W} = \sigma \mathbf{w}^{(1)} \circ \cdots \circ \mathbf{w}^{(N)}$, where $\sigma \geq 0$, $\| \mathbf{w}^{(n)}\|_1 = 1$, and the factors $\mathbf{w}^{(n)}, n=1, \cdots, N$ are identifiable up to sign flipping. Let $\mathbf{y} = [y^1, \cdots, y^M] \in \mathbb{R}^M $, and $\mathcal{X} = [\mathcal{X}^1, \cdots, \mathcal{X}^M] \in \mathbb{R}^{I_1 \times \cdots \times I_N \times M}$. Then \eqref{eq:MLR_SURT} can be reformulated as 
\begin{align}
   & \underset{\sigma, \mathbf{w}^{(n)}}{\min} \frac{1}{M} \| \mathbf{y} - \sigma \mathcal{X} \times_{1} \mathbf{w}^{(1)} \times_2 \cdots \times_N  \mathbf{w}^{(N)}\|_2^2 + \lambda \sigma \prod\nolimits_{n = 1}^N \|\mathbf{w}^{(n)}\|_1 + \alpha \sigma^2 \prod\nolimits_{n = 1}^N \|\mathbf{w}^{(n)}\|_2^2 \nonumber \\
 & \quad \text{s.t.} \quad \sigma \geq 0, ~ \| \mathbf{w}^{(n)} \|_1 = 1, ~ n= 1, \cdots, N.
  \label{eq:MLR_object_k}
\end{align}
Before diving into the stagewise/boosting algorithm, we first consider an alternating convex search (ACS) approach \citep{chen2012reduced,minasian2014energy} which appears to be natural for solving (\ref{eq:MLR_SURT}) with any fixed tuning parameter. Specifically, we alternately optimize with respect to a block of variables $(\sigma, \mathbf{w}^{(n)})$ with others fixed.
For each block $(\sigma, \mathbf{w}^{(n)})$, the relevant constraints are $\sigma \geq 0$ and $\| \mathbf{w}^{(n)} \|_1 = 1$, but the objective function in (\ref{eq:MLR_object_k}) is a function of $(\sigma, \mathbf{w}^{(n)})$ only through their product $\sigma \mathbf{w}^{(n)}$. So both constraints are avoided when optimizing with respect to $\sigma \mathbf{w}^{(n)}$. Let $\widehat{\mathbf{w}}^{(n)} = \sigma \mathbf{w}^{(n)}$ and $\mathbf{Z}^{(-n)} = \mathcal{X} \times_{1} \mathbf{w}^{(1)} \times_2 \cdots \times_{n-1} \mathbf{w}^{(n-1)} \times_{n+1} \cdots \times_N  \mathbf{w}^{(N)}$, the subproblem boils down to
% For simplicity, we drop the index $k$ from all the variables except $\mathbf{y}_k$ in the following.
% Our goal is to compute the entire solution path for $\mathcal{W}$, 
\begin{align}
   & \underset{\widehat{\mathbf{w}}^{(n)}}{\min} \frac{1}{M} \| \mathbf{y}^{\mathrm{T}} - \mathbf{Z}^{(-n)\mathrm{T}} \widehat{\mathbf{w}}^{(n)}\|_2^2 + \alpha \beta^{(-n)} \|\widehat{\mathbf{w}}^{(n)} \|_2^2 + \lambda \|\widehat{\mathbf{w}}^{(n)}\|_1, 
  \label{eq:MLR_rank_1_matrix_w}
\end{align}
where $\beta^{(-n)} = \prod_{l \neq n} \|\mathbf{w}^{(l)}\|_2^2$. Once we obtain the solution $\widehat{\mathbf{w}}^{(n)}$, we can set $\sigma = \| \widehat{\mathbf{w}}^{(n)} \|_1$ and $\mathbf{w}^{(n)} = \widehat{\mathbf{w}}^{(n)}/\sigma$ to satisfy the constraints whenever $\widehat{\mathbf{w}}^{(n)} \neq \mathbf{0}$. If $\widehat{\mathbf{w}}^{(n)} = \mathbf{0}$, $\mathbf{w}^{(n)}$ is no longer identifiable, we then set $\sigma=0$ and terminate the algorithm. Note that when $\alpha>0$, each subproblem is strongly convex and the generated solution sequence is uniformly bounded, we can show that ACS can converge to a coordinate-wise minimum point of \eqref{eq:MLR_object_k} with properly chosen initial value \citep{mishra2017sequential}. The optimization procedure then needs to be repeated to a grid of tuning parameter values for obtaining the solution paths of parameters and locating the optimal sparse solution along the paths.

%Although ACS is straightforward, it has several drawbacks. The optimization for each fixed tuning parameter 

%Moreover, since the problem is non-convex, these algorithms strongly depend on initialization and no strong convergence guarantees can be given \citep{da2015rank}. 

Inspired by the biconvex structure of (\ref{eq:MLR_SURT}) and the stagewise algorithm for LASSO \citep{zhao2007stagewise,VaughanChenYan2017}, we develop a fast stagewise unit-rank tensor factorization (SURF) algorithm to trace out the entire solution paths of (\ref{eq:MLR_SURT}) in a single run. The main idea of a stagewise procedure is to build a model from scratch, gradually increasing the model complexity in a sequence of simple learning steps. For instance, in stagewise estimation for linear regression, a forward step searches for the best predictor to explain the current residual and updates its coefficient by a small increment, and a backward step, on the contrary, may decrease the coefficient of a selected predictor to correct for greedy selection whenever necessary. Due to the biconvex or multi-convex structure of our objective function, it turns out that efficient stagewise estimation remains possible: the only catch is that, when we determine which coefficient to update at each iteration, we always get $N$ competing proposals from $N$ different tensor modes, rather than just one proposal in case of LASSO.

To simplify the notations, the objective function (\ref{eq:MLR_rank_1_matrix_w}) is re-arranged into a standard LASSO as 
\begin{align}
   & \underset{\widehat{\mathbf{w}}^{(n)}}{\min} \frac{1}{M} \| \widehat{\mathbf{y}} - \widehat{\mathbf{Z}}^{(-n)} \widehat{\mathbf{w}}^{(n)}\|_2^2 + \lambda \|\widehat{\mathbf{w}}^{(n)}\|_1
  \label{eq:MLR_compact}
\end{align}
using the augmented data 
$\widehat{\mathbf{y}} = (\mathbf{y}, \mathbf{0})^{\mathrm{T}}$ and $\widehat{\mathbf{Z}}^{(-n)} =  (\mathbf{Z}^{(-n)}, \sqrt{\alpha \beta^{(-n)}M} \mathbf{I})^{\mathrm{T}}$
%$\widehat{\mathbf{y}} = \begin{pmatrix} \mathbf{y} \\ \mathbf{0} \end{pmatrix}$ 
%and $\widehat{\mathbf{Z}}^{(-n)} = \begin{pmatrix} \mathbf{Z}^{(-n)\mathrm{T}} \\ \sqrt{\alpha \beta^{(-n)}M} \mathbf{I} \end{pmatrix} $
, where $\mathbf{I}$ is the identity matrix of size $I_n$. % but with the key difference being that the number of factors is not fixed, and new factors are incorporated in a greedy fashion.
We write the objective function (\ref{eq:MLR_compact}) by $$
\Gamma(\widehat{\mathbf{w}}^{(n)}; \lambda) = J(\widehat{\mathbf{w}}^{(n)}) + \lambda \Omega(\widehat{\mathbf{w}}^{(n)}).
$$
% when all variables are fixed but $\widehat{\mathbf{w}}^{(n)}$.
We use $(\sigma, \{\mathbf{w}^{(n)}\})$ to denote all the variables if necessary.
% We use var (q) to denote all the variables appearing in q and avar (q) to denote all its distinguished variables. We often abuse
% notation and use q to refer to the set of its atoms, ie,{α1,..., αm} …

\begin{savenotes}
\begin{algorithm}[!t]
\small
\caption{Fast Stagewise Unit-Rank Tensor Factorization (SURF)}
\label{alg:stagewise}
\begin{algorithmic}[1]
\REQUIRE Training data $\mathcal{D}$, a small stepsize $\epsilon>0$ and a small tolerance parameter $\xi$ \footnote{As stated in \citep{zhao2007stagewise}, $\xi$ is implementation-specific but not necessarily a user parameter. In all experiments, we set $\xi = \epsilon^2/2$.}
\ENSURE Solution paths of $(\sigma, \{\mathbf{w}^{(n)}\})$.
\STATE \textit{Initialization}: take a forward step with $( \{\hat{i}_1, \cdots, \hat{i}_N \}, \hat{s}) = \underset{\{i_1, \cdots, i_N\}, s=\pm\epsilon}{\argmin} J(s\mathbf{1}_{i_1}, \mathbf{1}_{i_2}, \cdots, \mathbf{1}_{i_N})$, and
\begin{align}
    \sigma_0 = \epsilon,  \mathbf{w}^{(1)}_0 = sign(\widehat{s})\mathbf{1}_{\widehat{i}_n},~ \mathbf{w}^{(n)}_0 = \mathbf{1}_{\widehat{i}_{n}} (n \neq 1),~\lambda_0 = (J(\{\mathbf{0}\}) - J(\sigma_0, \{\mathbf{w}^{(n)}_0\}))/\epsilon. \label{eq:initial}
\end{align}
% And calculate the initial regularization parameter
% \begin{equation}
% \lambda_0 = \frac{J(\{\mathbf{0}\}) - J(\hat{\mathbf{w}}^{(1)}_0, \{\mathbf{w}^{(-1)}_0\})}{\epsilon}.
% \end{equation}
Set the active index sets $I^{(n)}_0 = \{ \hat{i}_n\}$ for $n=1, \cdots, N$; $t=0$.
\REPEAT
\STATE \textit{Backward step}: %Find the backward step that leads to the most minimal loss for $n=1, \cdots, N$:
\begin{align}
    (\widehat{n}, \widehat{i}_{\widehat{n}}) := \arg\underset{n,i_n \in I^{(n)}_t}{\min}~ J(\widehat{\mathbf{w}}_t^{(n)} + s_{i_n}\mathbf{1}_{i_n}),~~\text{where}~~s_{i_n} = -sign(\widehat{w}_{ti_n}^{(n)})\epsilon.
    \label{eq:backward}
\end{align}
% \begin{numcases}{}
% \hspace{-1.5 mm} \hat{i}_n := \underset{i_n \in I^{(n)}_t}{\argmin}~ J(\hat{\mathbf{w}}^{(n)}_t + s_{i_n}\mathbf{1}_{i_n}; \lambda_t) \nonumber \\
%     \hspace{-1.5 mm} \hat{n} := \underset{n}{\argmin}~J(\hat{\mathbf{w}}^{(n)}_t + s_{\hat{i}_n} \mathbf{1}_{\hat{i}_n}; \lambda_t) \label{eq:backward}
% \end{numcases}
    \textbf{if} $\Gamma(\widehat{\mathbf{w}}^{(\widehat{n})}_t + s_{\widehat{i}_{\widehat{n}}}\mathbf{1}_{\widehat{i}_{\widehat{n}}}; \lambda_t) - \Gamma(\widehat{\mathbf{w}}^{(\widehat{n})}_t; \lambda_{t}) \leq - \xi$, \textbf{then}
    \begin{align}
    & \sigma_{t+1} = \|\widehat{\mathbf{w}}^{(\widehat{n})}_t + s_{\widehat{i}_{\widehat{n}}}\mathbf{1}_{\widehat{i}_{\widehat{n}}}\|_1,~\mathbf{w}^{(\widehat{n})}_{t+1}= (\widehat{\mathbf{w}}^{(\widehat{n})}_t + s_{\widehat{i}_{\widehat{n}}})/\sigma_{t+1}, ~\mathbf{w}^{(-\widehat{n})}_{t+1} = \mathbf{w}^{(-\widehat{n})}_{t}, \nonumber \\
       &\lambda_{t+1} = \lambda_t, ~~~~ I^{(n)}_{t+1} :=
        \begin{cases}
       I^{(\widehat{n})}_{t} \setminus \{\widehat{i}_{\widehat{n}}\} , & \text{if}~~w_{(t+1)\widehat{i}_{\widehat{n}}}^{(\widehat{n})} = 0 \\
       I^{(n)}_{t}, & \text{otherwise}. \nonumber
        \end{cases}
    \end{align}
\STATE \textbf{else} \textit{Forward step}: 
        \begin{align}
            & (\widehat{n}, \widehat{i}_{\widehat{n}}, \widehat{s}_{\widehat{i}_{\widehat{n}}}) := \arg\underset{n,i_n, s=\pm\epsilon}{\min}~J(\widehat{\mathbf{w}}^{(n)}_t + s_{i_n}\mathbf{1}_{i_n}), \label{eq:forward}\\
        & \sigma_{t+1} = \|\widehat{\mathbf{w}}^{(\widehat{n})}_t + \widehat{s}_{\widehat{i}_{\widehat{n}}}\mathbf{1}_{\widehat{i}_{\widehat{n}}}\|_1,~\mathbf{w}^{(\widehat{n})}_{t+1}= (\widehat{\mathbf{w}}^{(\widehat{n})}_t + \widehat{s}_{\widehat{i}_{\widehat{n}}})/\sigma_{t+1}, ~\mathbf{w}^{(-\widehat{n})}_{t+1} = \mathbf{w}^{(-\widehat{n})}_{t}, \nonumber \\
        & \lambda_{t+1} = \min[\lambda_t, \frac{J(\sigma_t, \{\mathbf{w}^{(n)}_t\}) - J(\sigma_{t+1}, \{\mathbf{w}^{(n)}_{t+1}\}) - \xi }{\Omega(\sigma_{t+1}, \{\mathbf{w}^{(n)}_{t+1}\}) - \Omega(\sigma_{t}, \{\mathbf{w}^{(n)}_{t}\}) }],~~~
         I^{(n)}_{t+1} :=
        \begin{cases}
            I^{(n)}_{t} \cup \{\widehat{i}_n \} , & \text{if}~~n = \widehat{n} \nonumber \\
            I^{(n)}_{t}, & \text{otherwise}. \nonumber
        \end{cases}
    \end{align} \vspace{-5mm}
    \STATE Set $t = t+1$.
\UNTIL{$\lambda_t \leq 0$}
\end{algorithmic}
\end{algorithm}
\end{savenotes}

The structure of our stagewise algorithm is presented in \textbf{Algorithm \ref{alg:stagewise}}. It can be viewed as a boosting procedure that builds up the solution gradually in terms of forward step (line 4) and backward step (line 3)\footnote{Boosting amounts to combine a set of ``weak learners'' to build one strong learner, and is connected to stagewise and regularized estimation methods. SURF is a boosting method since each backward/forward step is in essence finding a weaker learner to incrementally improve the current learner (thus generate a path of solutions).}. The initialization step is solved explicitly (see Lemma \ref{lemma:lambda_max} below). At each subsequent iteration, the parameter update takes the form $\widehat{\mathbf{w}}^{(n)} = \widehat{\mathbf{w}}^{(n)} + s\mathbf{1}_{i_n}$ in either forward or backward direction, where $\mathbf{1}_{i_n}$ is a length-$I_n$ vector with all zeros except for a $1$ in the $i_n$-th coordinate, $s=\pm\epsilon$, and $\epsilon$ is the pre-specified step size controlling the fineness of the grid. The algorithm also keeps track of the tuning parameter $\lambda$. Intuitively, the selection of the index $(n, i_n)$ and the increment $s$ is guided by minimizing the penalized loss function with the current $\lambda$ subject to a constraint on the step size. Comparing to the standard stagewise LASSO, the main difference here is that we need to select the ``best'' triple of $(n, i_n, s)$ over all the dimensions across all tensor modes.%; as such, our approach is a novel generalization of stagewise LASSO to tensor regression. 

Problem \eqref{eq:backward} and \eqref{eq:forward} can be solved efficiently. By expansion of $J(\widehat{\mathbf{w}}^{(n)} + s\mathbf{1}_{i_n})$, we have
\begin{align}
    J(\widehat{\mathbf{w}}^{(n)} + s\mathbf{1}_{i_n})= \frac{1}{M}(\|\widehat{\mathbf{e}}^{(n)}\|_2^2 - 2s \widehat{\mathbf{e}}^{(n)\mathrm{T}} \widehat{\mathbf{Z}}^{(-n)} \mathbf{1}_{i_n} + \epsilon^2 \|\widehat{\mathbf{Z}}^{(-n)} \mathbf{1}_{i_n}\|_2^2), \nonumber
\end{align}
where $\widehat{\mathbf{e}}^{(n)} = \widehat{\mathbf{y}} - \widehat{\mathbf{Z}}^{(-n)} \widehat{\mathbf{w}}^{(n)}$ is a constant at each iteration. Then the solution at each forward step is 
% \begin{align}
%     \hat{i}_n = \underset{i_n }{\argmin}~\frac{\epsilon}{2} \|\Delta_{i_n}\|_2^2-|\mathbf{e}_k^{\mathrm{T}}\Delta_{i_n}-2\alpha\beta^{(-n)}\hat{w}_{i_n}^{(n)}|,~\hat{s}= sign(\mathbf{e}_k^{\mathrm{T}}\Delta_{i_n}-2\alpha\beta^{(-n)}\hat{w}_{i_n}^{(n)}) \epsilon. % \label{eq:forward_opti}
% \end{align}
\begin{align}
    (\widehat{n}, \widehat{i}_{\widehat{n}}) := \underset{n, i_n }{\argmax}~2|\widehat{\mathbf{e}}^{(n)\mathrm{T}}\widehat{\mathbf{Z}}^{(-n)}\mathbf{1}_{i_n}| - \epsilon Diag(\widehat{\mathbf{Z}}^{(-n)\mathrm{T}}\widehat{\mathbf{Z}}^{(-n)})^{\mathrm{T}}\mathbf{1}_{i_n},~\widehat{s}= sign(\widehat{\mathbf{e}}^{(n)\mathrm{T}}\widehat{\mathbf{Z}}^{(-\widehat{n})}\mathbf{1}_{\widehat{i}_{\widehat{n}}}) \epsilon, \nonumber % \label{eq:forward_opti}
\end{align}
and at each backward step is
\begin{align}
    (\widehat{n}, \widehat{i}_{\widehat{n}}) := \underset{n, i_n \in I^{(n)} }{\argmin}~ 2sign(\widehat{w}_{i_n}^{(n)}) \widehat{\mathbf{e}}^{\mathrm{T}}\widehat{\mathbf{Z}}^{(-n)}\mathbf{1}_{i_n} + \epsilon Diag(\widehat{\mathbf{Z}}^{(-n)\mathrm{T}}\widehat{\mathbf{Z}}^{(-n)})^{\mathrm{T}}\mathbf{1}_{i_n}, \nonumber % \label{eq:forward_opti}
\end{align}
where $Diag(\cdot)$ denotes the vector formed by the diagonal elements of a square matrix, and $I^{(n)} \subset \{1, \cdots, I_n\}$ is the $n$-mode active index set at current iteration.

% The computational complexity of Algorithm \ref{alg:stagewise} is dominated by the matrix-vector multiplications involving $\hat{\mathbf{e}}^{\mathrm{T}}\hat{\mathbf{Z}}^{(-n)}$ and $\hat{\mathbf{Z}}^{(-n)\mathrm{T}}\hat{\mathbf{Z}}^{(-n)}$
\textbf{Computational Analysis}. In Algorithm \ref{alg:stagewise}, the most time-consuming part are the calculations of $\widehat{\mathbf{e}}^{(n)\mathrm{T}}\widehat{\mathbf{Z}}^{(-n)}$ and $Diag(\widehat{\mathbf{Z}}^{(-n)\mathrm{T}}\widehat{\mathbf{Z}}^{(-n)})$, involved in both forward and backward steps. We further write as $\widehat{\mathbf{e}}^{(n)\mathrm{T}}\widehat{\mathbf{Z}}^{(-n)} = (\mathbf{Z}^{(-n)}\mathbf{e} - \alpha \beta^{(-n)}M\widehat{\mathbf{w}}^{(n)})^{\mathrm{T}}$, $Diag(\widehat{\mathbf{Z}}^{(-n)\mathrm{T}}\widehat{\mathbf{Z}}^{(-n)})= Diag(\mathbf{Z}^{(-n)}\mathbf{Z}^{(-n)\mathrm{T}}) + \alpha\beta^{(-n)}M \mathbf{1}^{(n)}$, where $\mathbf{e} = \mathbf{y}^{\mathrm{T}} - \mathbf{Z}^{(-n) \mathrm{T}}\hat{\mathbf{w}}^{(n)}$ is a constant during each iteration but varies from one iteration to the next, $\mathbf{1}^{(n)}$ is a length-$I_n$ vector with all ones. At each iteration, the computational cost is dominated by the update of $\mathbf{Z}^{(-n)}$ ($n \neq \hat{n}$), which can be obtained by
\begin{equation}
    \mathbf{Z}^{(-n)}_{t+1} = \frac{1}{\sigma_{t+1} } (\sigma_{t} \mathbf{Z}^{(-n)}_{t}+ \mathbf{Z}^{(-n, -\hat{n})}_{t} \times_{\hat{n}} \widehat{s}_{\widehat{i}_{\widehat{n}}}\mathbf{1}_{\widehat{i}_{\widehat{n}}}),
\end{equation}
where $(-n, -\hat{n})$ denotes every mode except $n$ and $\hat{n}$, which greatly reduces the computation. Therefore, when $n\neq \hat{n}$, the updates of $\mathbf{Z}^{(-n)}$, $\mathbf{e}$, $\mathbf{Z}^{(-n)}\mathbf{e}$ and $Diag(\mathbf{Z}^{(-n)\mathrm{T}}\mathbf{Z}^{(-n)})$ requires $O(M\prod_{s \neq n, \hat{n}} I_s + 3M I_n)$, $O(M I_{\hat{n}})$, $O(M I_n)$, $O(M I_n)$ operations, respectively, when $n = \hat{n}$, we only need to additionally update $\mathbf{Z}^{(-\hat{n})}\mathbf{e}$, which requires $O(M I_{\hat{n}})$ operations. Overall the computational complexity of our approach is $O(M \sum_{n \neq \hat{n}}^N (\prod_{s \neq n, \hat{n}} I_s + 5I_n)+2M I_{\hat{n}})$ per iteration. In contrast, the ACS algorithm has to be run for each fixed $\lambda$, and within each of such problems it requires $O(M \prod_{n=1}^N I_n)$ per iteration \citep{da2015rank}.

\vspace{-5pt}
\section{Convergence Analysis}
\vspace{-5pt}
We provide convergence analysis for our stagewise algorithm in this section. All detailed proofs are given in the Appendix A. Specifically, Lemma \ref{lemma:lambda_max} and \ref{lemma:lambda_0} below justify the validity of the initialization.
% The following lemma determines $\lambda_{\max}$ and the initial non-zero solution of (\ref{eq:initial}) corresponding to $\lambda_{\max} - \epsilon$ explicitly.
\begin{lemma}\label{lemma:lambda_max}
Let $\mathbf{X}$ be the $(N+1)$-mode matricization of $\mathcal{X}$. Denote $\mathbf{X} = [\mathbf{x}_1, \cdots, \mathbf{x}_I]$ where each $\mathbf{x}_i$ is a column of $\mathbf{X}$, then
$$\lambda_{\max} = 2/M \max\{|\mathbf{x}_i^{\mathrm{T}} \mathbf{y}|; i = 1, \cdots,I.\}.$$
Moreover, letting $i^* = \arg\max_i |\mathbf{x}_i^{\mathrm{T}} \mathbf{y}|$ and $(i_1^*, \cdots, i_N^*)$ represents its corresponding indices in tensor space, then the initial non-zero solution of (\ref{eq:initial}), denoted as $(\sigma,\{\mathbf{w}^{(n)}\})$, is given by
$$\sigma=\epsilon, \mathbf{w}^{(1)} = sign(\mathbf{x}_{i^*}^{\mathrm{T}} \mathbf{y}) \mathbf{1}_{i_1^*},~\mathbf{w}^{(n)} = \mathbf{1}_{i_n^*}, \forall n=2, \cdots, N.$$
where $\mathbf{1}_{i_n^*}$ is a vector with all 0's except for a $1$ in the $i_n^*$-th coordinate.
\end{lemma}

\begin{lemma}\label{lemma:lambda_0}
If there exists $s$ and $i_n$ with $|s| =\epsilon, n=1, \cdots, N$ such that $\Gamma(s\mathbf{1}_{i_1}, \mathbf{1}_{i_2}, \cdots, \mathbf{1}_{i_N}; \lambda) \leq \Gamma(\{\mathbf{0}\}; \lambda)$, it must be true that $\lambda \leq \lambda_0$.
\end{lemma}
\vspace{-5pt}
Lemma \ref{lemma:backward_descent} shows that the backward step always performs coordinate descent update of fixed size $\epsilon$, each time along the steepest coordinate direction within the current active set, until the descent becomes impossible subject to a tolerance level $\xi$. Also, the forward step performs coordinate descent when $\lambda_{t+1}= \lambda_t$. Lemma \ref{lemma:update_descent} shows that when $\lambda$ gets changed, the penalized loss for the previous $\lambda$ can no longer be improved subject to a tolerance level $\xi$. Thus $\epsilon$ controls the granularity of the paths, and $\xi$ is a convergence threshold in optimizing the penalized loss with any fixed tuning parameter. They enable convenient trade-off between computational efficiency and estimation accuracy. 
\vspace{-5pt}
\begin{lemma}\label{lemma:backward_descent}
For any $t$ with $\lambda_{t+1}= \lambda_t$, we have $\Gamma(\sigma_{t+1}, \{\mathbf{w}^{(n)}_{t+1}\}; \lambda_{t+1}) \leq \Gamma(\sigma_t, \{\mathbf{w}^{(n)}_t\}; \lambda_{t+1}) - \xi$.
\end{lemma}
\vspace{-5pt}
\begin{lemma}\label{lemma:update_descent}
For any $t$ with $\lambda_{t+1} < \lambda_t$, we have $\Gamma(\hat{\mathbf{w}}^{(n)}_t + s_{i_n}\mathbf{1}_{i_n}; \lambda_t) > \Gamma(\hat{\mathbf{w}}^{(n)}_t; \lambda_t) - \xi.$
\end{lemma}
Lemma \ref{lemma:backward_descent} and Lemma \ref{lemma:update_descent} proves the following convergence theorem.
% \vspace{-5pt} 
\begin{exe}
% \begin{theorem}
\label{theorem:convergence}
For any $t$ such that $\lambda_{t+1} < \lambda_t$, we have $(\sigma_t, \{ \mathbf{w}^{(n)}_t\}) \rightarrow (\sigma(\lambda_t), \{\widetilde{\mathbf{w}}^{(n)}(\lambda_t)\})$ as $\epsilon, \xi \rightarrow 0$, where $(\sigma(\lambda_t), \{\widetilde{\mathbf{w}}^{(n)}(\lambda_t)\})$ denotes a coordinate-wise minimum point of Problem \eqref{eq:MLR_SURT}.
% \end{theorem}
\end{exe}
\vspace{-0.5pt}
\section{Experiments} \vspace{-0.5pt}
We evaluate the effectiveness and efficiency of our method \emph{\ours} through numerical experiments on both synthetic and real data, and compare with various state-of-the-art regression methods, including \emph{LASSO}, Elastic Net (\emph{ENet}), Regularized multilinear regression and selection (\emph{Remurs}) \citep{song2017multilinear}, optimal CP-rank Tensor Ridge Regression (\emph{orTRR}) \citep{guo2012tensor}, Generalized Linear Tensor Regression Model (\emph{GLTRM}) \citep{zhou2013tensor}, and a variant of our method with Alternating Convex Search (\emph{ACS}) estimation. Table \ref{tab:compared_methods} summarizes the properties of all methods. All methods are implemented in MATLAB and executed on a machine with 3.50GHz CPU and 256GB RAM. For LASSO and ENet we use the MATLAB package glmnet from \citep{friedman2010regularization}. For GLTRM, we solve the regularized CP tensor regression simultaneously for all $R$ factors based on TensorReg toolbox \citep{zhou2013tensor}.
% We follow \citep{kampa2014sparse} to arrange the test, validation and training sets in the format of (1:1:4), and then combine the training and validation sets into one dataset, and use it to train models. 
We follow \citep{kampa2014sparse} to arrange the test and training sets in the ratio of 1:5. The hyperparameters of all methods are optimized using 5-fold cross validation on the training set, with range $\alpha \in \{0.1, 0.2, \cdots, 1\}$, $\lambda \in \{10^{-3}, 5 \times 10^{-3}, 10^{-2}, 5 \times 10^{-2}, \cdots, 5 \times 10^2, 10^3 \}$, and $R \in \{1, 2, \cdots, 50\}$. Specifically, for GLTRM, ACS, and {\ours}, we simply set $\alpha =1$. For LASSO, ENet and ACS, we generate a sequence of 100 values for $\lambda$ to cover the whole path. For fairness, the number of iterations for all compared methods are fixed to 100. All cases are run 50 times and the average results on the test set are reported. Our code is available at \url{https://github.com/LifangHe/SURF}.
% We follow \citep{kampa2014sparse} to arrange the test, validation and training sets in the format of (1:1:4), and then combine the training and validation sets into one dataset, and use it to train models. Hyperparameters for all methods are optimized via five-fold cross validation on the combined training set.

% ----------- Property of Compared Methods ----------
% Each representing a different strategy
% ---------------------------------------------------
\begin{table*}[t]
\centering
% \small
\caption{Compared methods. $\alpha$ and $\lambda$ are regularized parameters; $R$ is the CP rank.} 
% $N$ is the number of samples; $M$ is the tensor dimension; $p =\prod_{m=1}^M I_m$; $s = \sum_{m=1}^M I_m$.}
\label{tab:compared_methods}
\begin{adjustbox}{max width=1\linewidth}
\begin{tabular}{|l|c|c|c|c|c|c|c|}
\hline
Methods               & LASSO   &  ENet   & Remurs & orTRR & GLTRM & ACS & \ours  \\ \hline
Input Data Type  & Vector  & Vector  & Tensor & Tensor & Tensor & Tensor & Tensor \\
Regularization     & $\ell_1$ ($\mathbf{w}$) & $\ell_1/\ell_2$ ($\mathbf{w}$) & Nuclear/$\ell_1$ ($\mathcal{W}$) & $\ell_2$ ($\mathbf{W}^{(n)}$) & $\ell_1/\ell_2$ ($\mathbf{W}^{(n)}$) & $\ell_1/\ell_2$ ($\mathcal{W}_r$)  & $\ell_1/\ell_2$ ($\mathcal{W}_r$) \\
Rank Explored         & --- & --- & --- & Optimized  & Fixed & Increased & Increased \\
\hline
Hyperparameters & $\lambda$ & $\alpha, \lambda$ & $\lambda_1, \lambda_2$ & $\alpha, R$  & $\alpha, \lambda, R$ & $\alpha,\lambda, R$ & $\alpha,\lambda, R$ \\
% Time Complexity    & $O(Npq)$  & $O(Npq)$ & $O(Npst)$ & ACS & $O((NR^2L+MPR)t)$ & ACS & Ours \\
% Estimation  & CD  & CD & ADMM & ACS & ACS & ACS & Ours \\
\hline
\end{tabular}
\end{adjustbox}
\label{tab:methods}
% \vspace{-1em}
\end{table*}

\textbf{Synthetic Data}. We first use the synthetic data to examine the performance of our method in different scenarios, with varying step sizes, sparsity level, number of features as well as sample size. We generate the data as follows: $y= \langle \mathbf{X}, \mathbf{W}\rangle + \varepsilon$, where $\varepsilon$ is a random noise generated from $\mathcal{N}(0, 1)$, and $\mathbf{X} \in \mathbb{R}^{I \times I}$. We generate $M$ samples $\{\mathbf{X}^m\}_{m=1}^M$ from $\mathcal{N}(0, \Sigma)$, where $\Sigma$ is a covariance matrix, the correlation coefficient between features $x_{i,j}$ and $x_{p,q}$ is defined as $0.6^{\sqrt{(i-p)^2+(j-q)^2}}$. We generate the true support as $\mathbf{W} = \sum_{r=1}^R \sigma_r \mathbf{w}^{(1)}_r \circ \mathbf{w}^{(2)}_r$, where each $\mathbf{w}^{(n)}_r \in \mathbb{R}^{I}$, $n=1, 2$, is a column vector with $\mathcal{N}(0, 1)$  i.i.d. entries and normalized with $\ell_1$ norm, the scalars $\sigma_r$ are defined by $\sigma_r = 1/r $. To impose sparsity on $\mathbf{W}$, we set $S\%$ of its entries (chosen uniformly at random) to zero. When studying one of factors, other factors are fixed to $M=500$, $I=16$, $R=50$, $S=80$.

\begin{figure*}[tbh!]
% \vspace{-5pt}
    \begin{minipage}{.33\textwidth}
      \centering
      \includegraphics[width=1.0\linewidth]{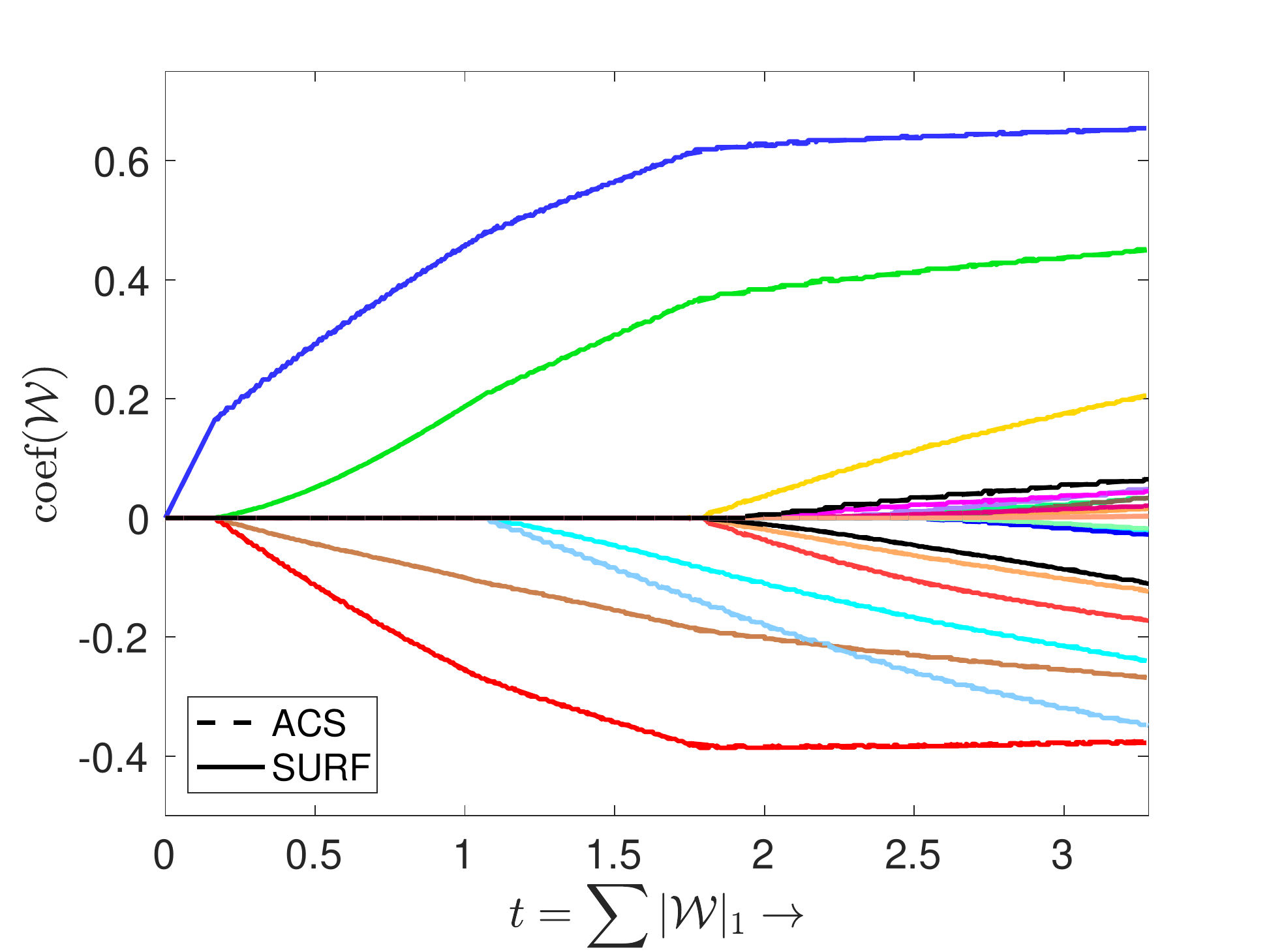}
      (a) $\epsilon = 0.01$
    %   \label{fig:path_01}
    \end{minipage}%
    \begin{minipage}{.33\textwidth}
      \centering
      \includegraphics[width=1.0\linewidth]{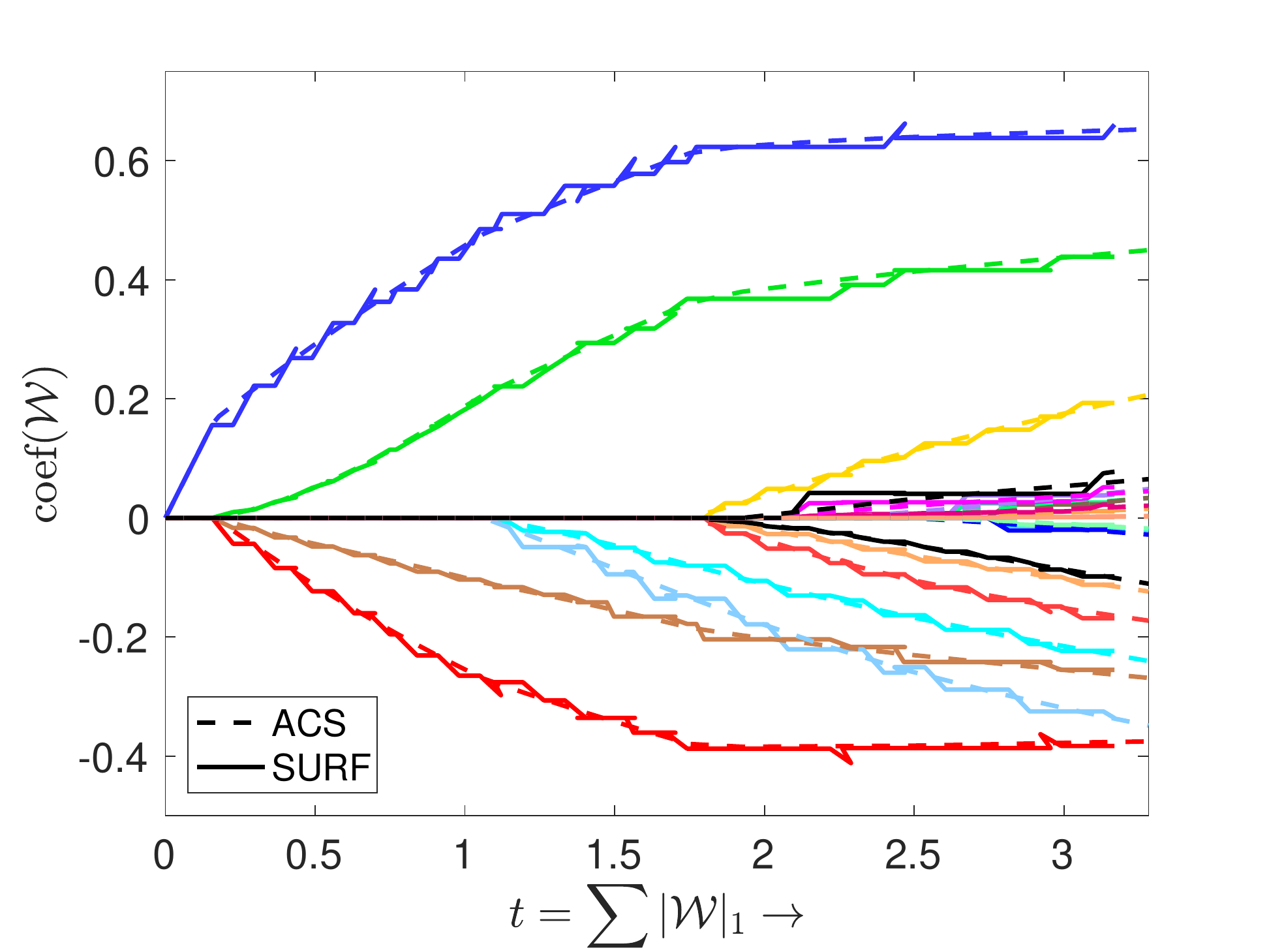}
      (b) $\epsilon = 0.1$
    %   \label{fig:path_1}
    \end{minipage}
    \begin{minipage}{.33\textwidth}
      \centering
      \includegraphics[width=1.0\linewidth]{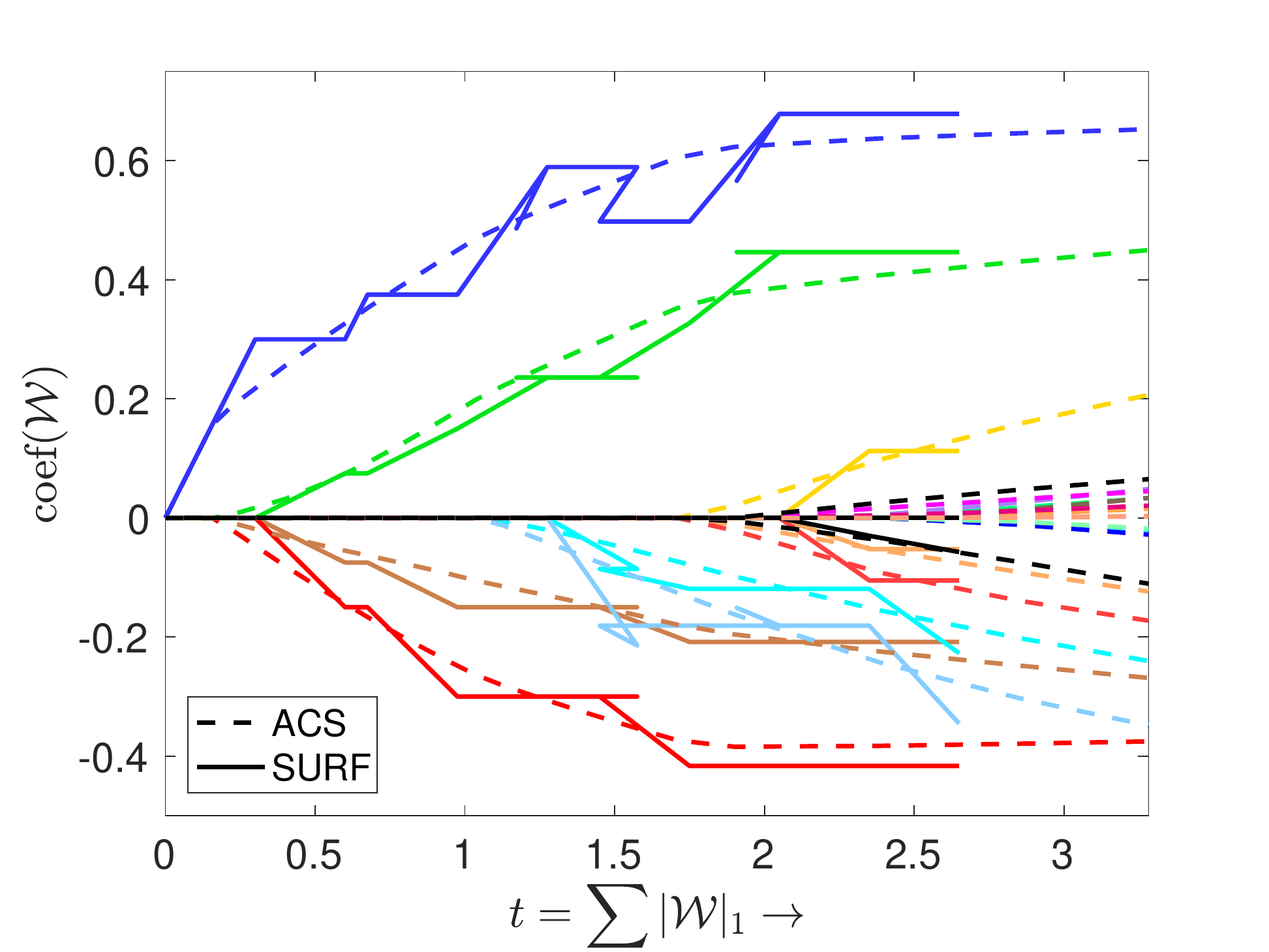}
      (c) $\epsilon = 0.5$
    %   \label{fig:path_5}
    \end{minipage}
\caption{\label{fig:solution_path} Comparison of solution paths of {\ours} (solid line) and ACS (dashed line) with different step sizes on synthetic data. The path of estimates $\mathcal{W}$ for each $\lambda$ is treated as a function of $t = \|\mathcal{W}\|_1$. Note that when the step size is small, the {\ours} path is almost indiscernible from the ACS path.}
% \vspace{-5pt}
\end{figure*}

\begin{wrapfigure}{r}{0.34\textwidth}
  \vspace{-15pt}
  \centering
    \includegraphics[width=0.34\textwidth]{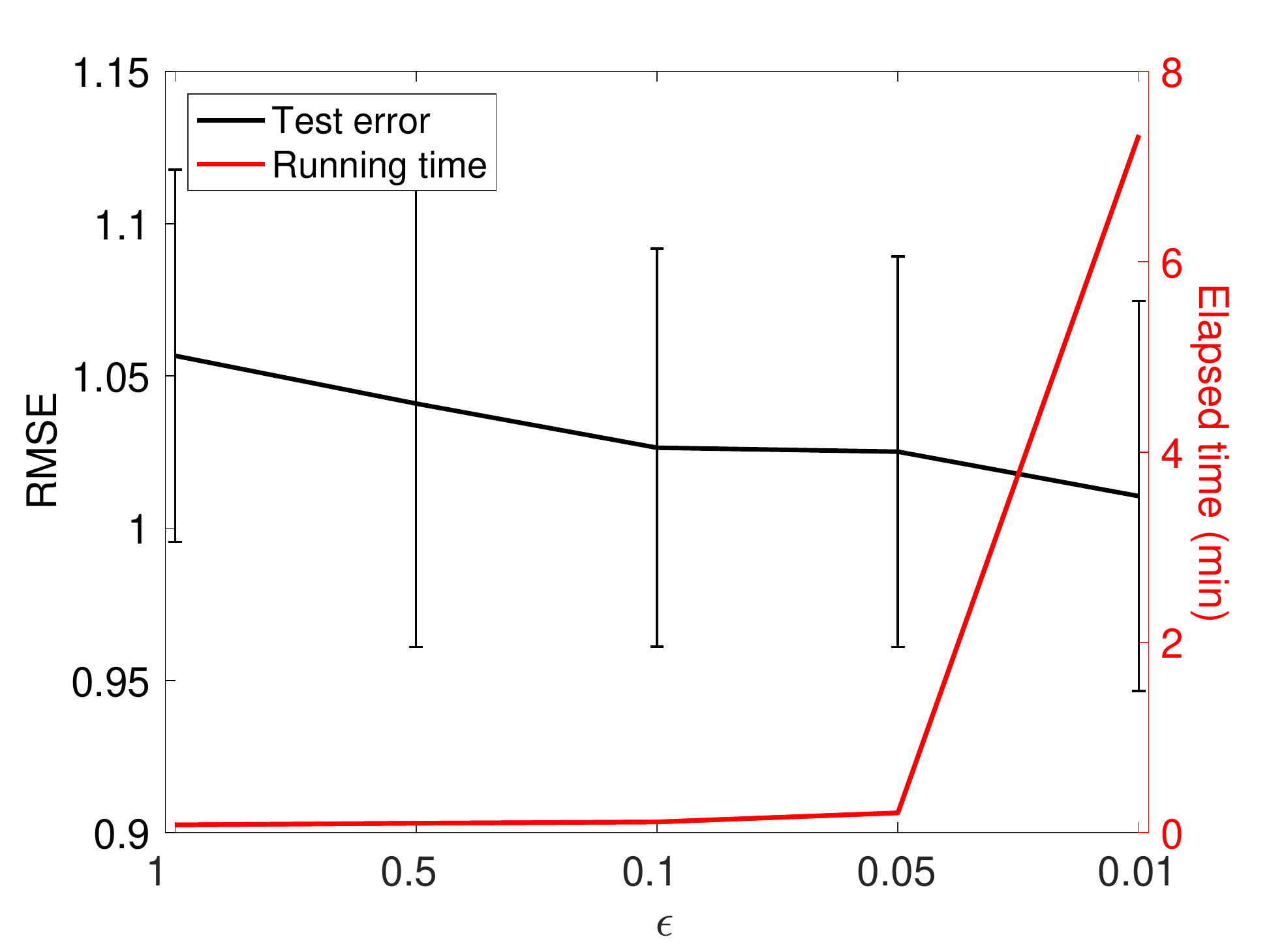}
  \vspace{-15pt}
  \caption{\label{fig:step_size} Results to different values of step size $\epsilon$.}
  \vspace{-15pt}
\end{wrapfigure}

% We provide a coefficient path plot of our method versus ACS method in Figure 1 
In a first step we analyze the critical parameter $\epsilon$ for our method. This parameter controls how close {\ours} approximates the ACS paths. Figure \ref{fig:solution_path} shows the solution path plot of our method versus ACS method under both big and small step sizes. As shown by the plots, a smaller step size leads to a closer approximation to the solutions of ACS. In Figure \ref{fig:step_size}, we also provide a plot of averaged prediction error with standard deviation bars (left side of y-axis) and CPU execution time (right side of y-axis in mins) over different values of step size. From the figure, we can see that the choice of the step size affects both computational speed and the root mean-squared prediction error (RMSE). The smaller the value of step size, the more accurate the regression result but the longer it will take to run. In both figures, the moderate step size $\epsilon = 0.1$ seems to offer a better trade-off between performance and ease of implementation. In the following we fix $\epsilon = 0.1$.

\begin{figure*}[tbh!]
\vspace{-5pt}
\centering
    \begin{minipage}{.33\textwidth}
      \centering
      \includegraphics[width=1.0\linewidth]{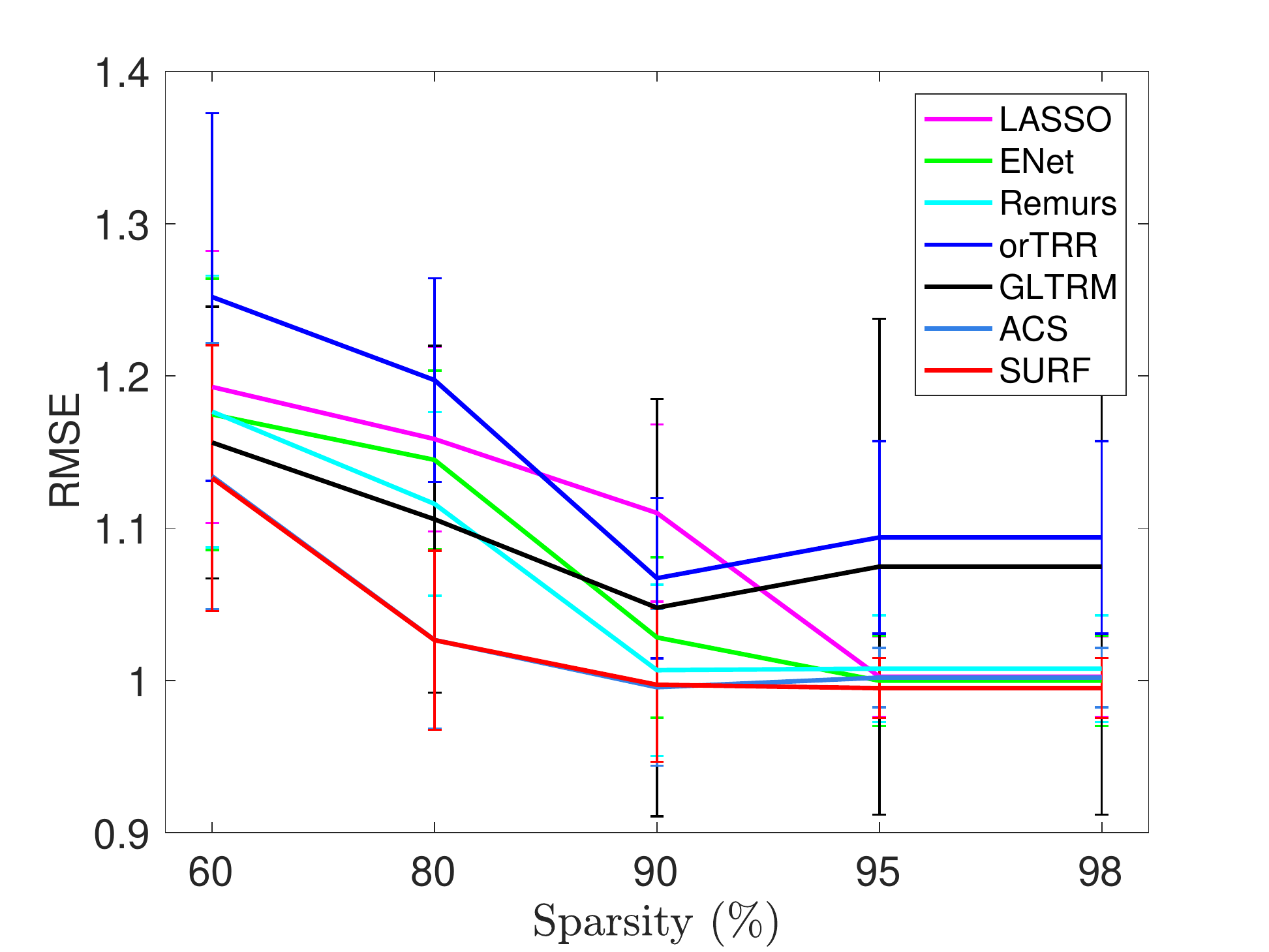}
      (a) Test error
      %\caption{(a)}
      %\label{fig:RMSE}
    \end{minipage}%
    \begin{minipage}{.33\textwidth}
      \centering
      \includegraphics[width=1.0\linewidth]{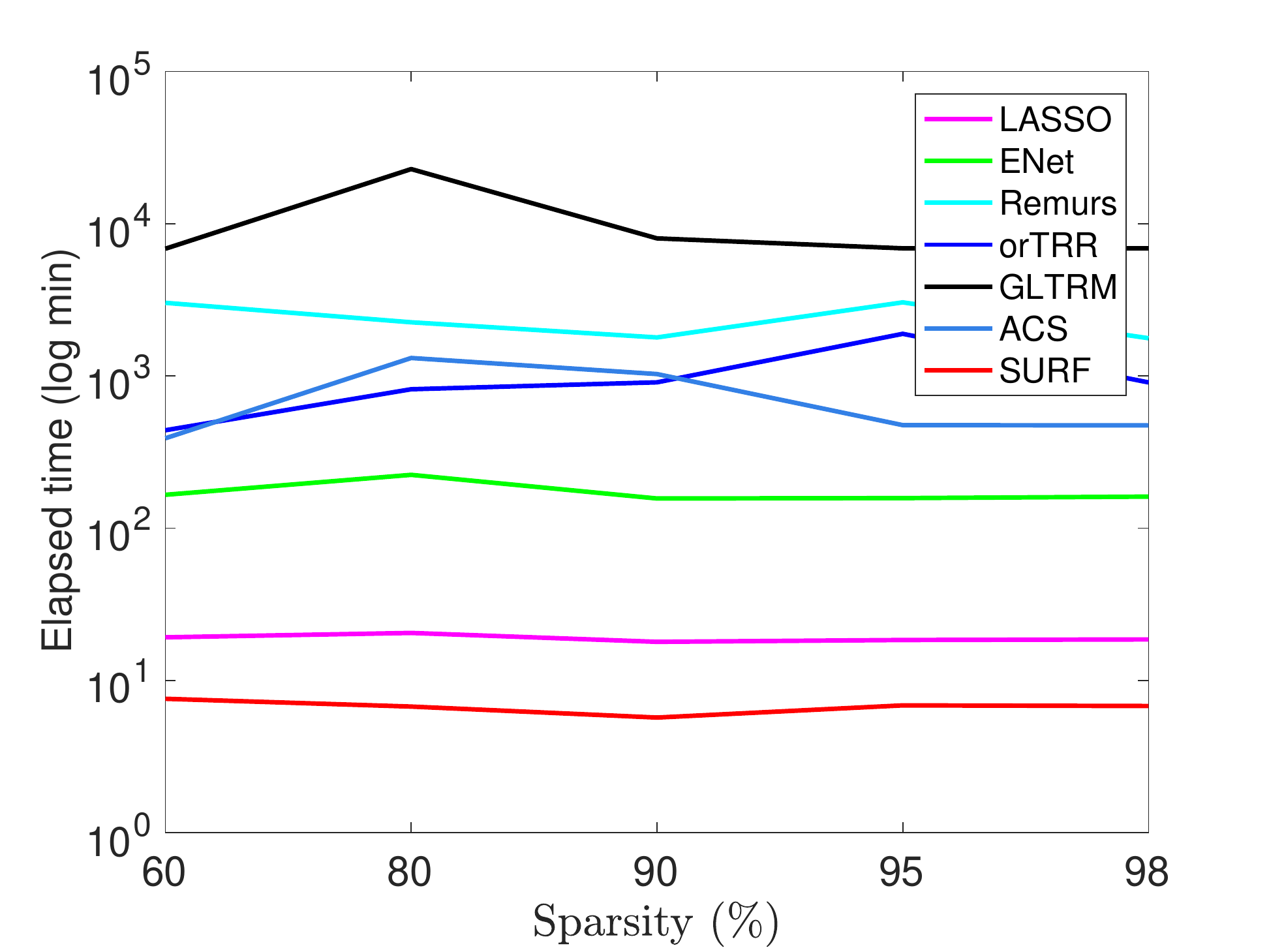}
      (b) Running time
      %\caption{(b)}
      %\label{fig:Time}
    \end{minipage}
    \begin{minipage}{.33\textwidth}
      \centering
      \includegraphics[width=1.05\linewidth]{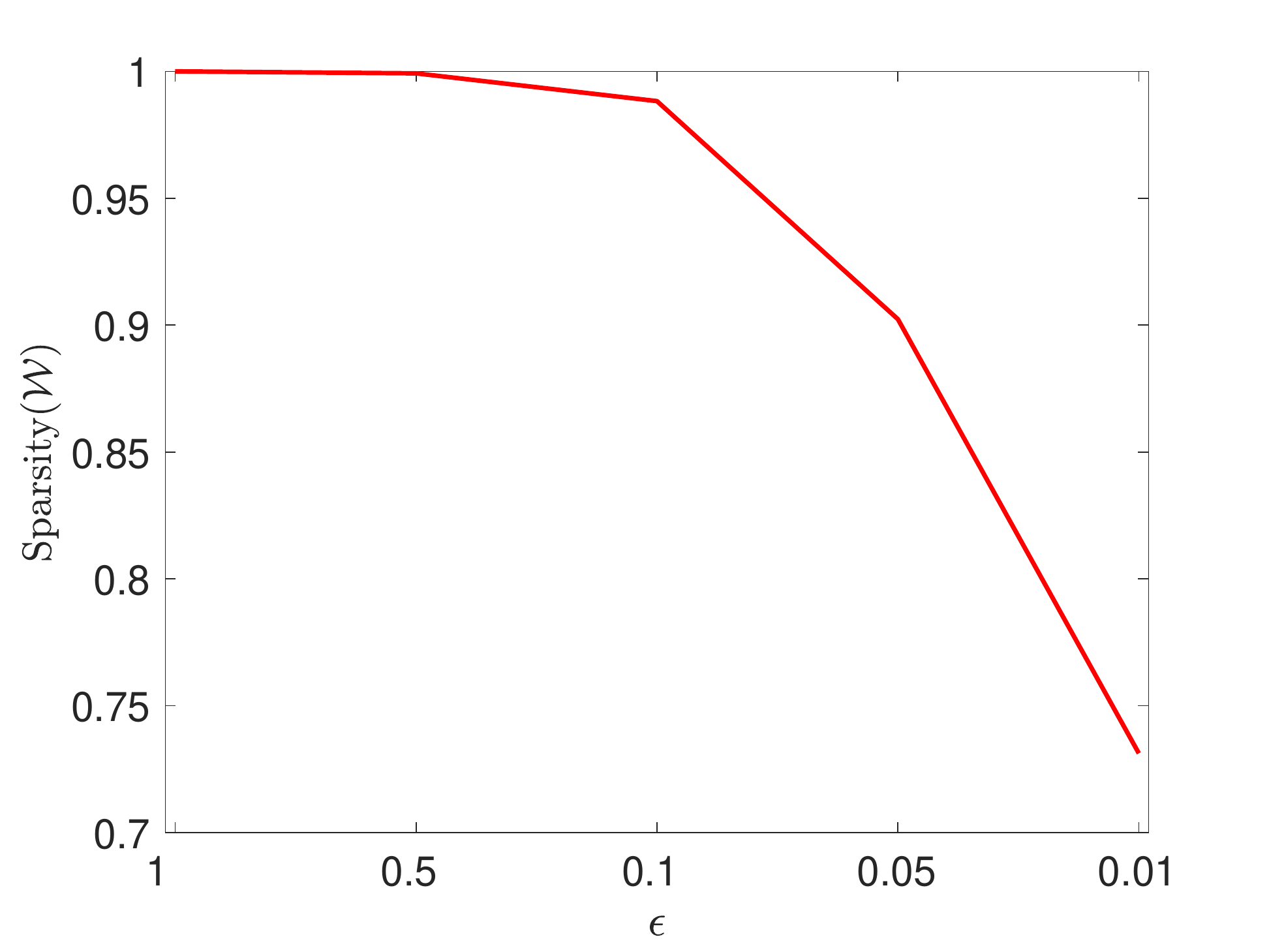}
      (c) Sparsity vs. $\epsilon$
      %\caption{(b)}
      %\label{fig:Time}
    \end{minipage}
    \vspace{-5pt}
\caption{\label{fig:differ_SC} Results with increasing sparsity level ($S\%$) of true $\mathbf{W}$ on synthetic 2D data (a)-(b), and (c) sparsity results of $\mathcal{W}$ versus step size for {\ours}.}
\end{figure*}

Next, we examine the performance of our method with varying sparsity level of $\mathbf{W}$. For this purpose, we compare the prediction error (RMSE) and running time (log min) of all methods on the synthetic data. Figure \ref{fig:differ_SC}(a)-(b) shows the results for the case of $S=\{60, 80, 90, 95, 98\}$ on synthetic 2D data, where $S\%$ indicates the sparsity level of true $\mathbf{W}$. As can be seen from the plots, {\ours} generates slightly better predictions than other existing methods when the true $\mathbf{W}$ is sparser. Moreover, as shown in Figure \ref{fig:differ_SC}(c), it is also interesting to note that larger step sizes give much more sparsity of coefficients for {\ours}, this explains why there is no value for some curves as the increase of $\sum|\mathcal{W}|_1$ in Figure \ref{fig:solution_path}(c). 

\begin{figure*}[tbh!]
\vspace{-5pt}
    \begin{minipage}{.33\textwidth}
      \centering
      \includegraphics[width=1.05\linewidth]{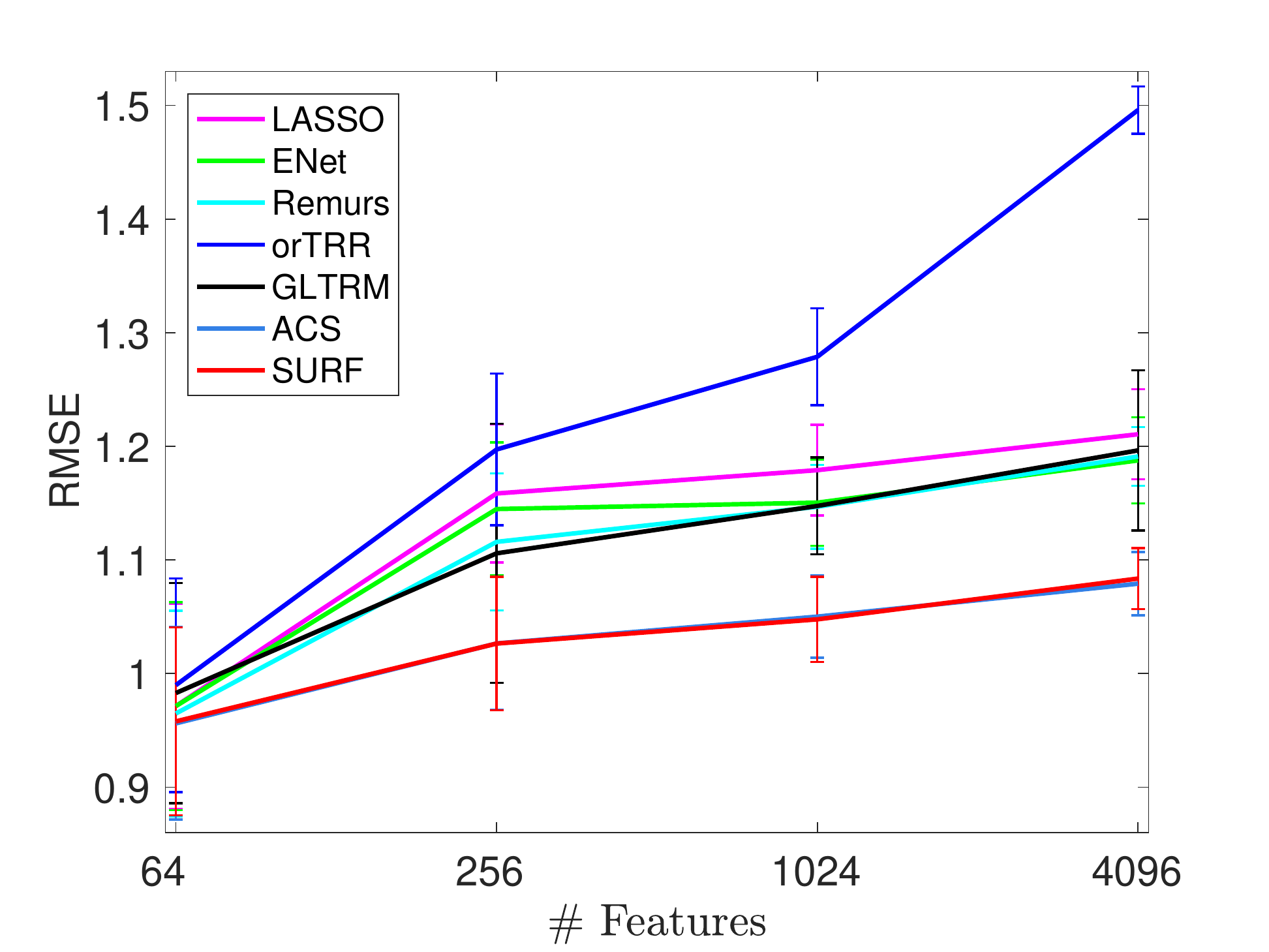}
      (a) Test error
      %\caption{(a)}
      %\label{fig:RMSE}
    \end{minipage}%
    \begin{minipage}{.33\textwidth}
      \centering
      \includegraphics[width=1.05\linewidth]{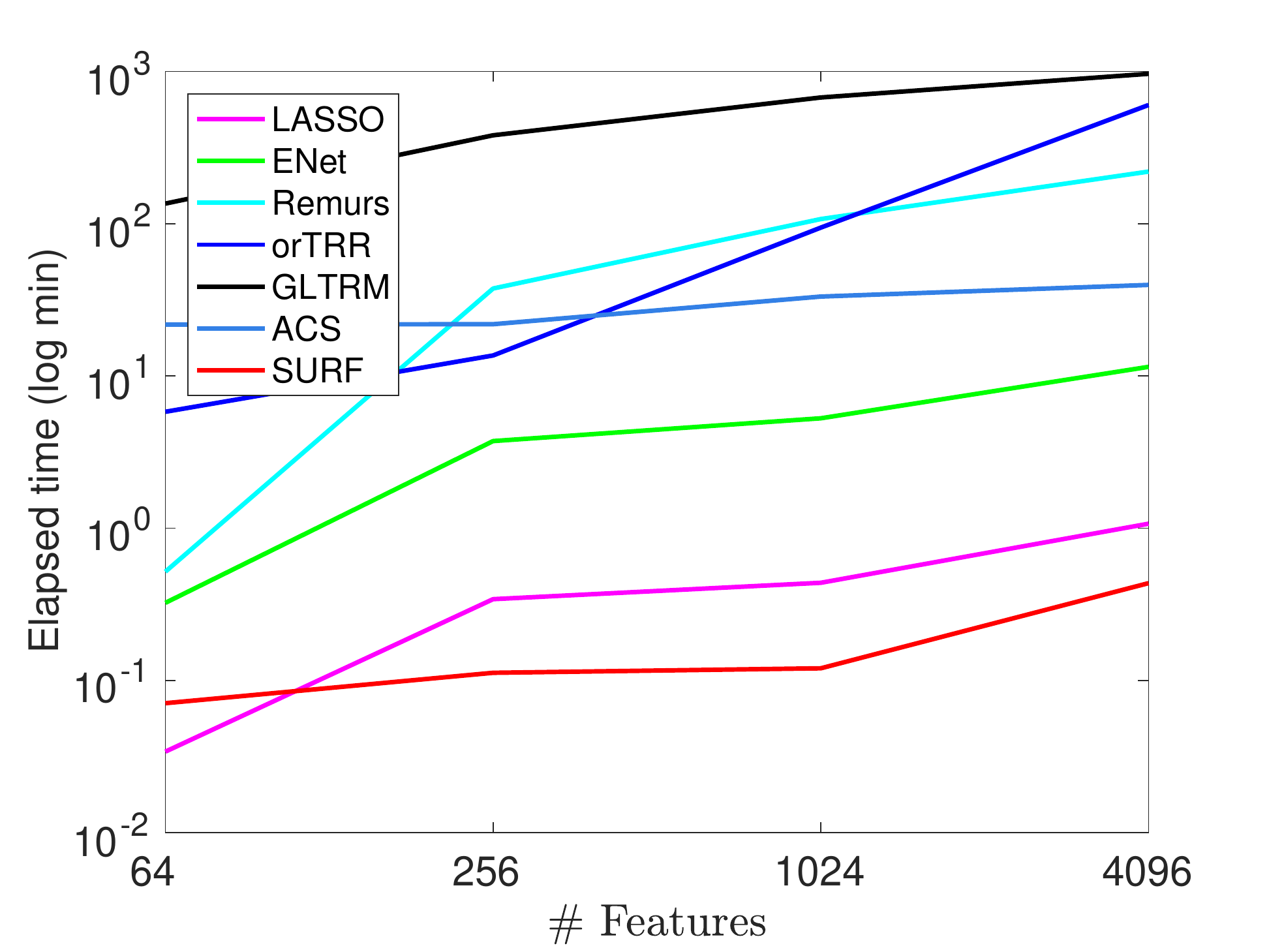}
      (b) Running time
      %\caption{(b)}
      %\label{fig:Time}
    \end{minipage}
    \begin{minipage}{.33\textwidth}
      \centering
      \includegraphics[width=1.05\linewidth]{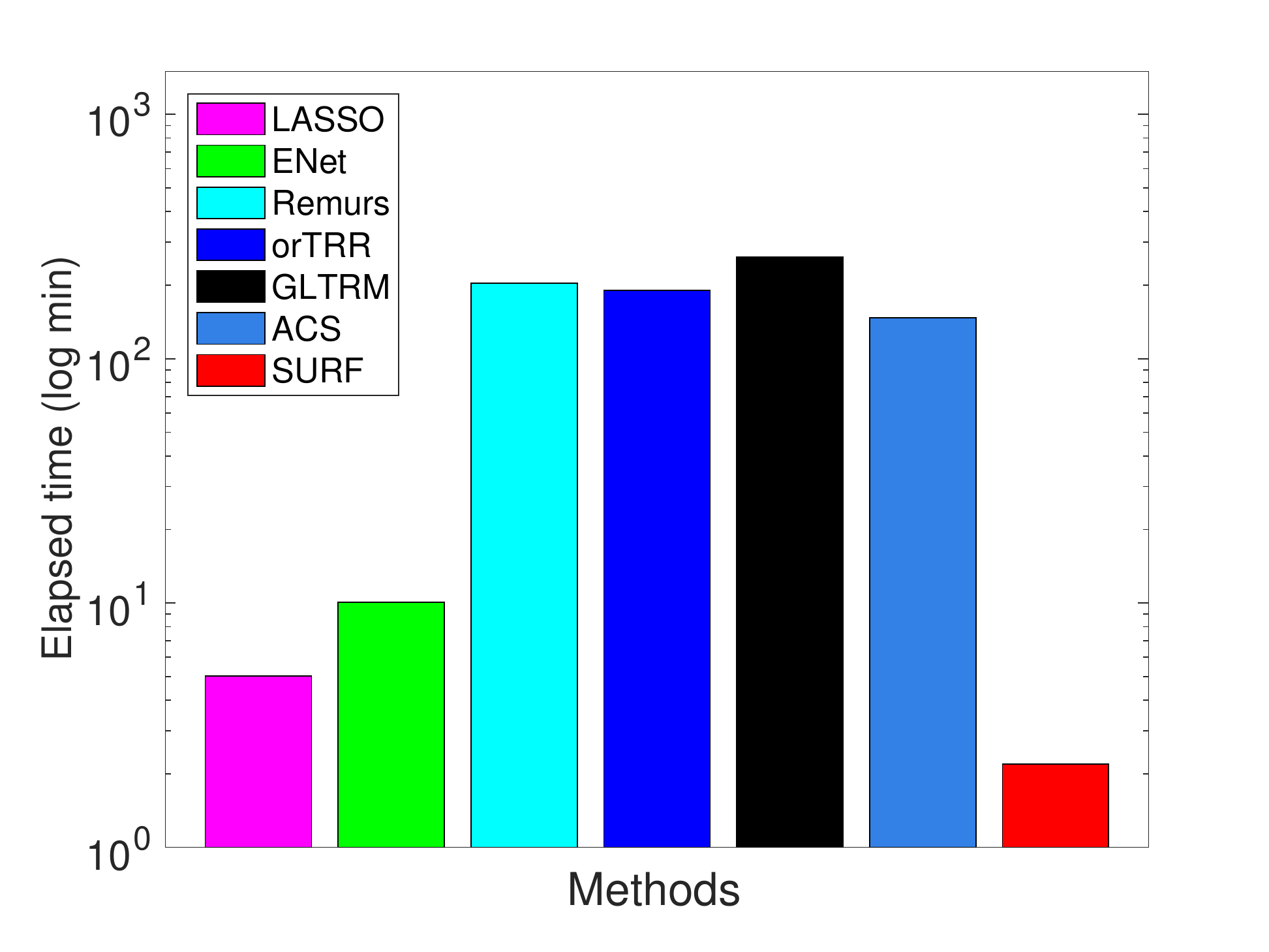}
      (c) Running time on MRI data
      %\caption{(b)}
      %\label{fig:Time}
    \end{minipage}
\caption{\label{fig:differ_feature} Results with increasing number of features on synthetic 2D data (a)-(b), and (c) real 3D MRI data of features $240 \times 175 \times 176$ with fixed hyperparameters (without
cross validation).}
\end{figure*}

Furthermore, we compare the prediction error and running time (log min) of all methods with increasing number of features. Figure \ref{fig:differ_feature}(a)-(b) shows the results for the case of $I=\{8, 16, 32, 64\}$ on synthetic 2D data. Overall, {\ours} gives better predictions at a lower computational cost. Particularly, {\ours} and ACS have very similar prediction qualities, this matches with our theoretical result on the solutions of {\ours} versus ACS. {\ours} achieves better predictions than other tensor methods, indicating the effectiveness of structured sparsity in unit-rank tensor decomposition itself. In terms of running time, it is clear that as the number of features is increased, {\ours} is significantly faster than other methods. 

\begin{figure*}[tbh!]
\vspace{-5pt}
\centering
    \begin{minipage}{.35\textwidth}
      \centering
      \includegraphics[width=1.0\linewidth]{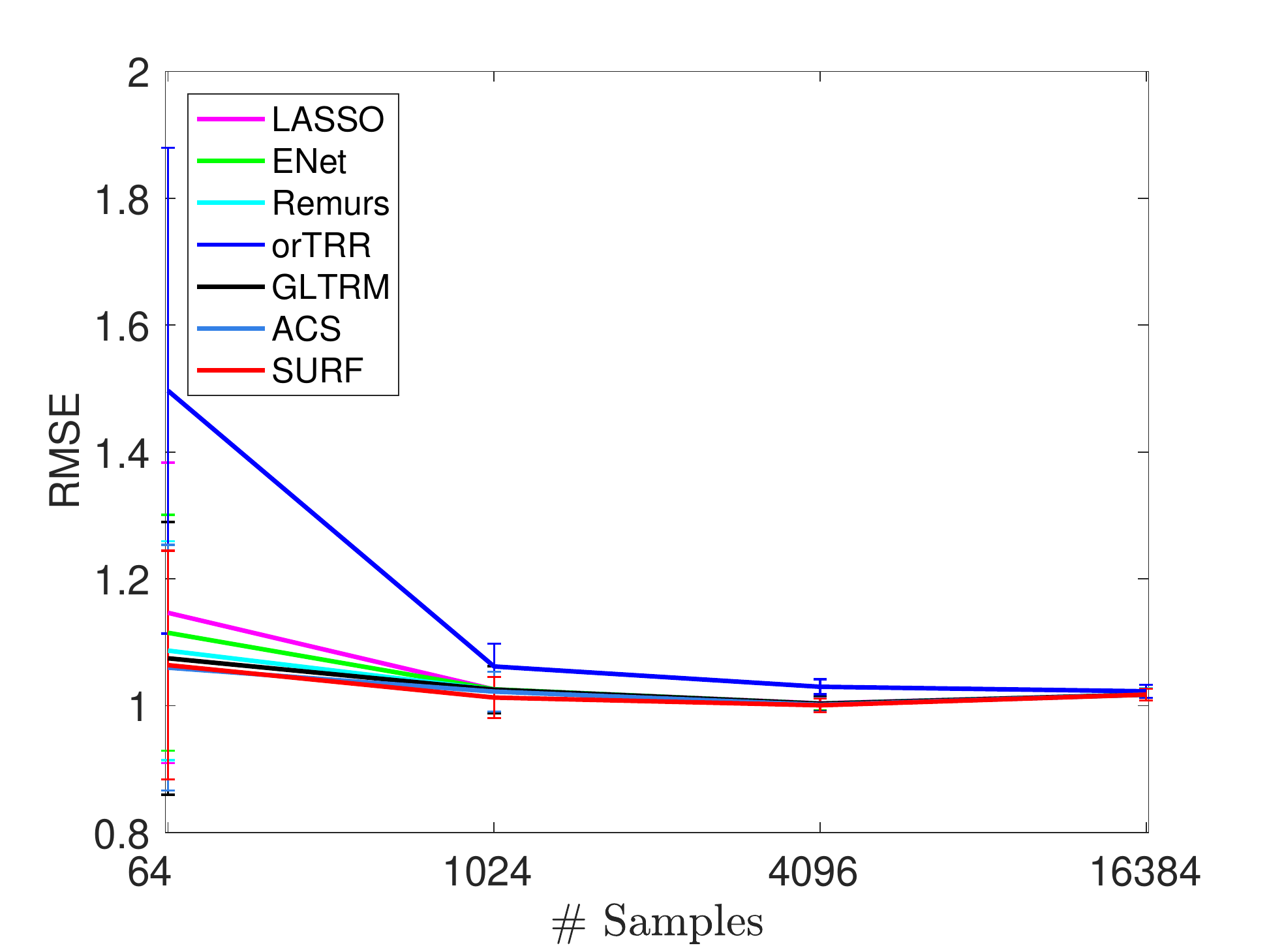}
      (a) Test error
      %\caption{(a)}
      %\label{fig:RMSE}
    \end{minipage}%
    \begin{minipage}{.35\textwidth}
      \centering
      \includegraphics[width=1.0\linewidth]{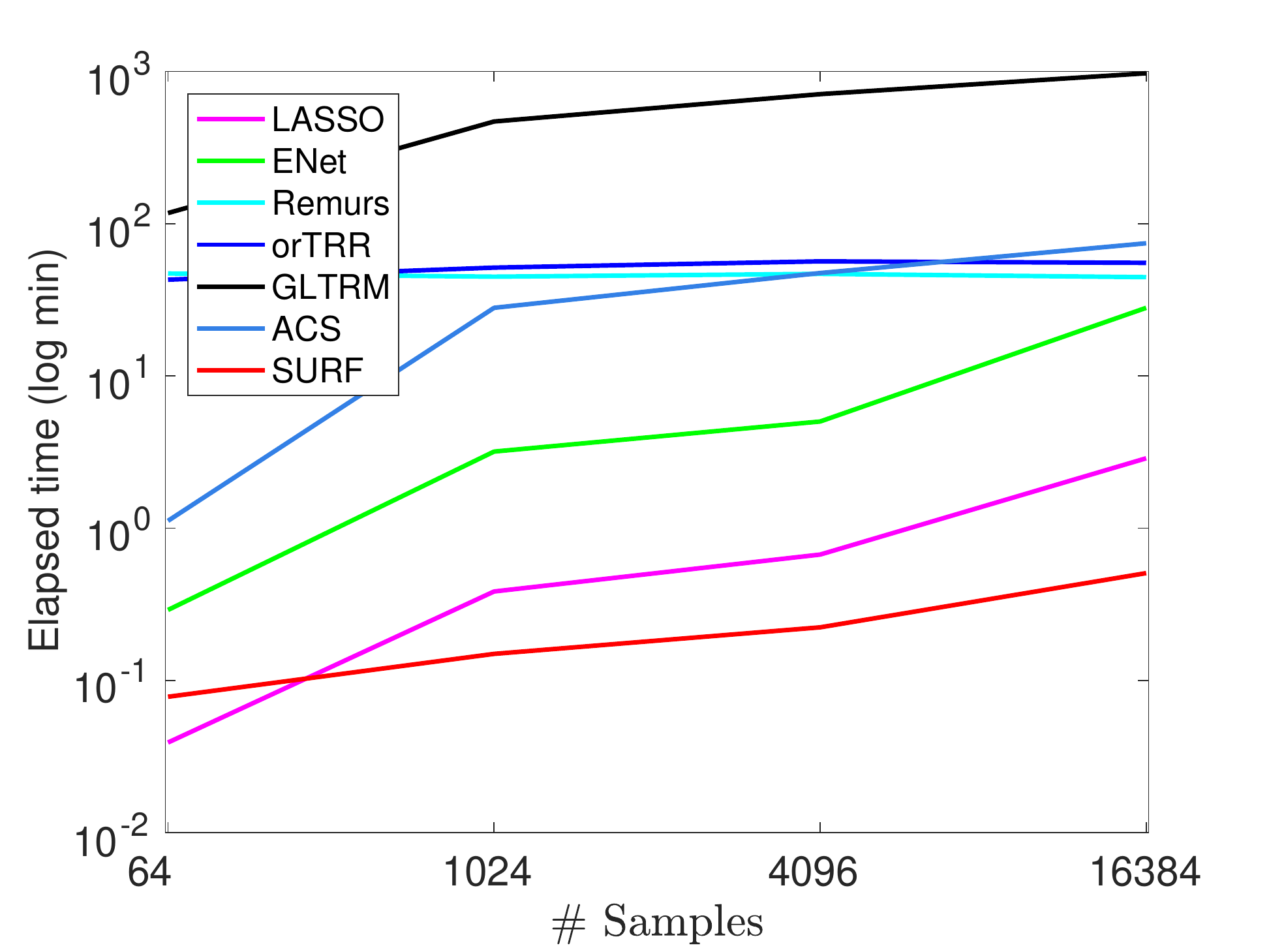}
      (b) Running time
      %\caption{(b)}
      %\label{fig:Time}
    \end{minipage}
    % \begin{minipage}{.33\textwidth}
    %   \centering
    %   \includegraphics[width=1.05\linewidth]{nips2018/figure/MRI_bar_box_nostd.pdf}
    %   (c) Running time on MRI data
    %   %\caption{(b)}
    %   %\label{fig:Time}
    % \end{minipage}
\caption{\label{fig:differ_sample} Results with increasing number of samples on synthetic 2D data.}
\end{figure*}

Finally, Figure \ref{fig:differ_sample} shows the results with increasing number of samples. Overall, {\ours} gives better predictions at a lower computational cost. Particularly, Remurs and orTRR do not change too much as increasing number of samples, this may due to early stop in the iteration process when searching for optimized solution.

\vspace{-0.5em}

% \begin{wrapfigure}{r}{0.33\textwidth}
%   \vspace{-20pt}
%   \begin{center}
%     \includegraphics[width=0.33\textwidth]{nips2018/figure/Sim_Sam_Time.pdf}
%   \end{center}
%   \vspace{-10pt}
%   \caption{\label{fig:sample_size_time} Results with increasing number of samples on synthetic 2D data.}
%   \vspace{-20pt}
% \end{wrapfigure}

\begin{table*}[h]
\centering
% \small
\scriptsize
\caption{Performance comparison over different DTI datasets. Column 2 indicates the used metrics RMSE, Sparsity of Coefficients (SC) and CPU execution time (in mins). The results are averaged over 50 random trials, with both the mean values and standard deviations (mean $\pm$ std.)}
\label{tab:exp_result}
\begin{tabular}{|c|c|c|c|c|c|c|c|c|}
\hline
\multirow{2}{*}{Datasets}     & \multirow{2}{*}{Metrics} & \multicolumn{7}{c|}{Comparative Methods}                                                                            \\ \cline{3-9} 
                              &                          & LASSO         & ENet          & Remurs          & orTRR         & GLTRM           & ACS             & {\ours}          \\ \hline
\multirow{3}{*}{DTI$_{fact}$} & RMSE                     & 2.94$\pm$0.34 & 2.92$\pm$0.32 & 2.91$\pm$0.32   & 3.48$\pm$0.21 & 3.09$\pm$0.35   & 2.81$\pm$0.24   & 2.81$\pm$0.23 \\ \cline{2-9} 
                              & Sparsity                 & 0.99$\pm$0.01 & 0.97$\pm$0.01 & 0.66$\pm$0.13   & 0.00$\pm$0.00 & 0.90$\pm$0.10   & 0.92$\pm$0.02   & 0.95$\pm$0.01 \\ \cline{2-9} 
                              & Time                     & 6.4$\pm$0.3   & 46.6$\pm$4.6  & 161.3$\pm$9.3   & 27.9$\pm$5.6  & 874.8$\pm$29.6  & 60.8$\pm$24.4   & 1.7$\pm$0.2   \\ \hline
\multirow{3}{*}{DTI$_{rk2}$}  & RMSE                     & 3.18$\pm$0.36 & 3.16$\pm$0.42 & 2.97$\pm$0.30   & 3.76$\pm$0.44 & 3.26$\pm$0.46   & 2.90$\pm$0.31   & 2.91$\pm$0.32 \\ \cline{2-9} 
                              & Sparsity                 & 0.99$\pm$0.01 & 0.95$\pm$0.03 & 0.37$\pm$0.09   & 0.00$\pm$0.00 & 0.91$\pm$0.06   & 0.93$\pm$0.02   & 0.94$\pm$0.01 \\ \cline{2-9} 
                              & Time                     & 5.7$\pm$0.3   & 42.4$\pm$2.9  & 155.0$\pm$10.7  & 10.2$\pm$0.1  & 857.4$\pm$22.5  & 63.0$\pm$21.6   & 5.2$\pm$0.8   \\ \hline
\multirow{3}{*}{DTI$_{sl}$}   & RMSE                     & 3.06$\pm$0.34 & 2.99$\pm$0.34 & 2.93$\pm$0.27   & 3.56$\pm$0.41 & 3.14$\pm$0.39   & 2.89$\pm$0.38   & 2.87$\pm$0.35 \\ \cline{2-9} 
                              & Sparsity                 & 0.98$\pm$0.01 & 0.95$\pm$0.01 & 0.43$\pm$0.17   & 0.00$\pm$0.00 & 0.87$\pm$0.03   & 0.90$\pm$0.03   & 0.93$\pm$0.02 \\ \cline{2-9} 
                              & Time                     & 5.8$\pm$0.3   & 45.0$\pm$1.0  & 163.6$\pm$9.0   & 7.5$\pm$0.9   & 815.4$\pm$6.5   & 66.3$\pm$44.9   & 1.5$\pm$0.1   \\ \hline
\multirow{3}{*}{DTI$_{tl}$}   & RMSE                     & 3.20$\pm$0.40 & 3.21$\pm$0.59 & 2.84$\pm$0.35   & 3.66$\pm$0.35 & 3.12$\pm$0.32   & 2.82$\pm$0.33   & 2.83$\pm$0.32 \\ \cline{2-9} 
                              & Sparsity                 & 0.99$\pm$0.01 & 0.96$\pm$0.03 & 0.44$\pm$0.13   & 0.00$\pm$0.00 & 0.86$\pm$0.03   & 0.90$\pm$0.02   & 0.91$\pm$0.02 \\ \cline{2-9} 
                              & Time                     & 5.5$\pm$0.2   & 42.3$\pm$1.4  & 159.6$\pm$7.6   & 26.6$\pm$3.1  & 835.8$\pm$9.9   & 96.7$\pm$43.2   & 3.8$\pm$0.5   \\ \hline
\multirow{3}{*}{Combined}     & RMSE                     & 3.02$\pm$0.37 & 2.89$\pm$0.41 & 2.81$\pm$0.31   & 3.33$\pm$0.27 & 3.26$\pm$0.45   & 2.79$\pm$0.31   & 2.78$\pm$0.29 \\ \cline{2-9} 
                              & Sparsity                 & 0.99$\pm$0.00 & 0.97$\pm$0.01 & 0.34$\pm$0.22   & 0.00$\pm$0.00 & 0.91$\pm$0.19   & 0.97$\pm$0.01   & 0.99$\pm$0.00 \\ \cline{2-9} 
                              & Time                     & 8.8$\pm$0.6   & 71.9$\pm$2.3  & 443.5$\pm$235.7 & 48.4$\pm$8.0  & 1093.6$\pm$49.7 & 463.2$\pm$268.1 & 6.2$\pm$0.5   \\ \hline
\end{tabular}
\vspace{-5pt}
\end{table*}

\textbf{Real Data}. We also examine the performance of our method on a real medical image dataset, including both DTI and MRI images, obtained from the Parkinson's progression markers initiative (PPMI) database\footnote{\url{http://www.ppmi-info.org/data }} with $656$ human subjects. We parcel the brain into 84 regions and extract four types connectivity matrices. Our goal is to predict the Montreal Cognitive Assessment (MoCA) scores for each subject. Details of data processing are presented in Appendix B. We use three metrics to evaluate the performance: root mean squared prediction error (RMSE), which describes the deviation between the ground truth of the response and the predicted values in out-of-sample testing; sparsity of coefficients (SC), which is the same as the $S\%$ defined in the synthetic data analysis (i.e., the percentage of zero entries in the corresponding coefficient); and CPU execution time. Table \ref{tab:exp_result} shows the results of all methods on both individual and combined datasets. Again, we can observe that {\ours} gives better predictions at a lower computational cost, as well as good sparsity. In particular, the paired t-tests showed that for all five real datasets, the RMSE and SC of our approach are significantly lower and higher than those of Remurs and GLTRM methods, respectively. This indicates that the performance gain of our approach over the other low-rank + sparse methods is indeed significant. Figure \ref{fig:differ_feature}(c) provides the running time (log min) of all methods on the PPMI MRI images of $240\times175\times176$ voxels each, which clearly demonstrates the efficiency of our approach.

\section*{Acknowledgements}
This work is supported by NSF No. IIS-1716432 (Wang), IIS-1750326 (Wang), IIS-1718798 (Chen), DMS-1613295 (Chen), IIS-1749940 (Zhou), IIS-1615597 (Zhou), ONR N00014-18-1-2585 (Wang), and N00014-17-1-2265 (Zhou), and Michael J. Fox Foundation grant number 14858 (Wang). Data used in the preparation of this article were obtained from the Parkinson's Progression Markers Initiative (PPMI) database (\url{http://www.ppmi-info.org/data}). For up-to-date information on the study, visit \url{http://www.ppmi-info.org}. PPMI -- a public-private partnership -- is funded by the Michael J. Fox Foundation for Parkinson's Research and funding partners, including Abbvie, Avid, Biogen, Bristol-Mayers Squibb, Covance, GE, Genentech, GlaxoSmithKline, Lilly, Lundbeck, Merk, Meso Scale Discovery, Pfizer, Piramal, Roche, Sanofi, Servier, TEVA, UCB and Golub Capital.

\small
\bibliographystyle{unsrtnat}
\bibliography{09-reference}

\end{document}